\documentclass[lettersize,journal]{IEEEtran}
\usepackage{amsmath,amsfonts}
\usepackage{algorithmic}
\usepackage{array}
\usepackage[caption=false,font=normalsize,labelfont=sf,textfont=sf]{subfig}
\usepackage{textcomp}
\usepackage{stfloats}
\usepackage{url}
\usepackage{verbatim}
\usepackage{graphicx}
\usepackage{mathrsfs}
\usepackage{amssymb}
\def\BibTeX{{\rm B\kern-.05em{\sc i\kern-.025em b}\kern-.08em
		T\kern-.1667em\lower.7ex\hbox{E}\kern-.125emX}}
\usepackage{balance}
\usepackage{multirow}
\newtheorem{theorem}{Theorem}
\usepackage{xcolor}
\begin{document}
	\title{Multichannel Orthogonal Transform-Based Perceptron Layers for Efficient ResNets}
	\author{Hongyi Pan,~\IEEEmembership{Member,~IEEE}, Emadeldeen Hamdan,~\IEEEmembership{Graduate Student Member,~IEEE}, Xin Zhu,~\IEEEmembership{Graduate Student Member,~IEEE}, Salih Atici, Ahmet Enis Cetin,~\IEEEmembership{Fellow,~IEEE}
		\thanks{This work was supported by National Science Foundation (NSF) grants 1934915, 2229659, U.S. Department of Energy (DOE) grant DE-SC0023715, and National Institutes of Health (NIH) grant R01-CA240639. The source code of this work is available at https://github.com/phy710/transform-based-layers.}}
	
	
	\maketitle
	
	\begin{abstract}
		In this paper, we propose a set of transform-based neural network layers as an alternative to the $3\times3$ Conv2D layers in Convolutional Neural Networks (CNNs). The proposed layers can be implemented based on orthogonal transforms such as the Discrete Cosine Transform (DCT), Hadamard transform (HT), and biorthogonal Block Wavelet Transform (BWT). Furthermore, by taking advantage of the convolution theorems, convolutional filtering operations are performed in the transform domain using element-wise multiplications. Trainable soft-thresholding layers, that remove noise in the transform domain, bring nonlinearity to the transform domain layers. Compared to the Conv2D layer, which is spatial-agnostic and channel-specific, the proposed layers are location-specific and channel-specific. Moreover, these proposed layers reduce the number of parameters and multiplications significantly while improving the accuracy results of regular ResNets on the ImageNet-1K classification task. Furthermore, they can be inserted with a batch normalization layer before the global average pooling layer in the conventional ResNets as an additional layer to improve classification accuracy. 
	\end{abstract}
	

	\section{Introduction}
	\IEEEPARstart{R}{ecent} literature states that convolutional neural networks (CNNs) have shown remarkable performance in many computer vision tasks, such as image classification~\cite{he2016deep, badawi2020computationally, agarwal2021coronet}, object detection~\cite{redmon2016you, aslan2020deep, menchetti2019pain, aslan2019early, beratouglu2021vehicle}, semantic segmentation~\cite{yu2018bisenet, huang2019ccnet, long2015fully, poudel2019fast, jin2019fast} and clustering~\cite{dong2017short, koyuncu2022centroidal, miao2022federated}. One of the most famous and successful CNNs is ResNet~\cite{he2016deep}, which enables the construction of very deep networks to get better performance in accuracy. However, with more and more layers being used, the network becomes bloated. In this case, the huge number of parameters increases the computational load for the devices, especially for the edge devices with limited computational resources. The convolutional layer, which is the critical component in the CNNs, is our target not only to slim but also to improve the accuracy of the network.
	
	In the convolutional layer, convolution kernels are spatial-agnostic and channel-specific. Because of the spatial-agnostic characteristic, a convolutional layer cannot adapt to different visual patterns with respect to different spatial locations. Therefore, in the convolutional layer, many redundant convolutional filters and parameters are required for feature extraction in location-related problems.
	
	The well-known Fourier convolution theorem states that the convolution in the space domain is equivalent to the element-wise multiplication in the Fourier domain. In other words, $\mathbf{y}=\mathbf{a}*\mathbf{x}$ in the space domain can be implemented in the Discrete Fourier Transform (DFT) domain by element-wise multiplication: 
	\begin{equation}
		\mathbf{Y}[k] = \mathbf{A}[k]\mathbf{X}[k], \label{eq: DFT Convolution}
	\end{equation}
	where $\mathbf{Y}= \mathcal{F}(\mathbf{y}), \mathbf{A}= \mathcal{F}(\mathbf{a})$, and $\mathbf{X}= \mathcal{F}(\mathbf{x})$ are the DFTs of $\mathbf{y}, \mathbf{a}$, and $\mathbf{x}$, respectively, and $\mathcal{F}(\cdot)$ is the Fourier transform operator.
	Eq.~(\ref{eq: DFT Convolution}) holds when the size of the DFT is larger than the size of the convolution output $\mathbf{y}$. Since there is a one-to-one relationship between the kernel coefficients $\mathbf{a}$ and the Fourier transform representation $\mathbf{A}$,     
	kernel weights can be learned in the Fourier transform domain using backpropagation-type algorithms. As the DFT is an orthogonal ($\mathbf{F}_N^H \mathbf{F}_N= \mathbf{F}_N\mathbf{F}_N^H =\mathbf{I}$, where $\mathbf{F}_N$ is the $N$ by $N$ transform matrix),
	and complex-valued transform, we consider other orthogonal transforms such as the Discrete Cosine Transform (DCT) and the Hadamard transform (HT). DCT is a real-valued transform and HT is a binary transform that is multiplication-free. In this paper, we use the DCT and HT convolution theorems to develop novel network layers. Other orthogonal transforms such as the orthogonal transforms generated from wavelet packet filter banks~\cite{cetin1993block} and the discrete Hartley transform~\cite{hartley1942more} can be also used in our approach.
	Furthermore, we implement the block wavelet transform (BWT) using the Wavelet biorthogonal 1.3 filters (Bior 1.3) to develop novel network layers. 
	
	
	Other related transform domain methods include:
	
	\noindent {\bf DFT-based methods} The Fast Fourier transform (FFT) algorithm is the most important signal and image processing method. As it is well-known Eq.~(\ref{eq: DFT Convolution}), convolutions in time and image domains can be performed using elementwise multiplications in the Fourier domain. However, the Discrete Fourier Transform (DFT) is a complex transform. In~\cite{chi2020fast, mohammad2021substitution}, the fast Fourier Convolution (FFC) method is proposed in which the authors designed the FFC layer based on the so-called Real-valued Fast Fourier transform (RFFT). In RFFT-based methods, they concatenate the real and the imaginary parts of the FFT outputs, and they perform convolutions in the transform domain with ReLU in the concatenated Fourier domain. Therefore, they do not take advantage of the Fourier domain convolution theorem. Moreover, concatenating the real and the imaginary parts of the complex tensors increases the number of parameters so their model requires more parameters than the original ResNets as the number of channels is doubled after concatenating. On the contrary, our method takes advantage of the convolution theorem and it can reduce the number of parameters significantly while producing comparable and even higher accuracy results. Moreover, in~\cite{rao2023gfnet}, Rao~\textit{et al. propose a global filter using DFT. They also apply element-wise multiplication in the transform for the vision transformers. However, as it is mentioned above, DFT is a complex transform, which increases the computational cost. On the other hand, our proposed transform domain methods do not require complex arithmetic. Furthermore, we perform soft-thresholding denoising in the transform domain. The soft-thresholding nonlinearity was proposed for denoising in 1995~\cite{david1995denoising} for transform domain signal and image denoising methods and we used it as a part of a deep neural network.
	}
	
	\noindent {\bf DCT-based methods} Other DCT-based methods include~\cite{gueguen2018faster, dos2020good, dos2021less, xu2021dct, ulicny2022harmonic}. Since the images are stored in the DCT domain in JPEG format, the authors in~\cite{gueguen2018faster, dos2020good, dos2021less} use DCT coefficients of the JPEG images but they did not take the transform domain convolution theorem to train the network. In~\cite{xu2021dct}, transform domain convolutional operations are used only during the testing phase. The authors did not train the kernels of the CNN in the DCT domain.  
	They only take advantage of the fast DCT computation and reduce parameters by changing $3\times 3$ Conv2D layers to $2\times 2$. In contrast, we train the filter parameters in the DCT domain and we use the soft-thresholding operator as the nonlinearity. We train a network using DCT with such an antisymmetric function because both positive and negative entries in the DCT domain are equally important. 
	In~\cite{ulicny2022harmonic}, Harmonic convolutional networks based on DCT were proposed. Only forward DCT without inverse DCT computation is employed to obtain Harmonic blocks for feature extraction. In contrast, with spatial convolution with learned kernels, this study proposes feature learning by weighted combinations of responses of predefined filters. The latter extracts harmonics from lower-level features in a region, and it applies DCT on the outputs from the previous layer which are already encoded by DCT.
	
	\noindent {\bf HT-Based Neural Networks} 
	In~\cite{zhao2021zero}, authors use HT to assist their ``ZerO Initialization" method. They apply the HT in the skip connections. However, they did not use the convolution theorem and they did not replace the convolutional layers with multiplicative layers in the Hadamard domain. This method does not reduce the number of parameters or the computational cost. Their goal is to improve the accuracy of their networks. HT-based neural networks from our early work including~\cite{deveci2018energy, pan2021fast,pan2022block} are used to reduce the computational cost. In~\cite{deveci2018energy}, a binary neural network with a two-stream structure is proposed, where one input is the regular image and the other is the HT of the image, and convolutions are not implemented in the transform domain. The HT is only applied at the beginning in~\cite{deveci2018energy}. The HT-based layers in~\cite{pan2021fast,pan2022block} also take the element-wise multiplication in the transform domain like this work, but they do not extract any channel-wise feature. They only extract the features width-wise and height-wise, as they have a similar scaling layer to this work. On the contrary, we have channel-wise processing layers in the transform domain to extract the channel-wise features, and this revision improves the performance of the HT-based layer significantly.

	\noindent {\bf Wavelet-based methods} Wavelet transform (WT) is a well-established method in signal and image processing. WT-based neural networks include~\cite{liu2019multi,oyallon2018compressing}. However, the regular WT does not have a convolution theorem. In other words, convolutions cannot be equivalently implemented in the wavelet domain. Instead, we convert the wavelet transform to block wavelet transform (BWT) as in \cite{cetin1993block}. Additionally, we assign weights to the BWT coefficients as in convolution theorems and use them in a network layer as if we are performing convolutions in the transform domain; \textit{i.e.}, we perform elementwise multiplications in the transform domain instead of convolutions. Authors in ~\cite{liu2019multi,oyallon2018compressing} did not take advantage of the BWT.
	
	
	\noindent {\bf Trainable soft-thresholding} The soft-thresholding function is commonly used in wavelet transform domain denoising~\cite{david1995denoising}. It is a proximal operator for the $\ell_1$ norm~\cite{karakucs2020simulation}. With trainable threshold parameters, soft-thresholding and its variants can be employed as the nonlinear function in the frequency domain-based networks~\cite{pan2021fast,pan2022block,badawi2021discrete,pan2022deep}. As a result of the importance of the positive and negative values in the transform domain, ReLU is not suitable. A completely positive waveform can have both significant positive and negative values in the transform domain. Lacking the negative-valued frequency components may cause issues when the inverse transform is applied. The soft-thresholding function keeps both positive and negative valued frequency components whose magnitudes exceed the trainable threshold, and its computational cost is similar to the one in ReLU.
	
	Our contribution is summarized as follows:
	
	\noindent $\bullet$ We propose a family of orthogonal transform domain approaches to replace the convolutional layer in a CNN to reduce parameters and computational costs. With the proposed layers, a CNN can reach a comparable or even higher accuracy with significantly fewer parameters and less computational consumption.
	
	\noindent $\bullet$ We extend the proposed structures to biorthogonal wavelet transforms. Both orthogonal such as the Haar wavelet and biorthogonal such as Bior 1.3 can be used to construct Conv2D layers.
	
	\noindent $\bullet$ We propose multi-channel structures for the proposed layers. In this paper, we use the tri-channel structure. This tri-channel design contains more parameters and MACs than the single-channel layer, but it makes the revised CNNs reach higher accuracy results. For example, compared to the regular ResNet-50 model, ResNet-50 with the tri-channel DCT-perceptron layer reaches 0.82\% higher center-crop top-1 accuracy on the ImageNet-1K dataset with 11.5\% fewer parameters and 11.5\% fewer MACs. 
	
	\noindent $\bullet$ The proposed single-channel layer can be inserted before the global average pooling layer or the flattened layer to improve the accuracy of the network. For instance, without changing the convolutional base, an extra single-channel DCT-perceptron layer improves 
	the center-crop top-1 accuracy of ResNet-18 0.74\% on the ImageNet-1K dataset with only 2.3\% extra parameters and 0.3\% extra MACs.

	\section{Methodology}
	\subsection{Orthogonal Transforms and Their Convolution Theorems}
	Like the discrete Fourier transform (DFT), the discrete cosine transform (DCT) is also widely used for frequency analysis because it can be considered a real-valued version of the DFT~\cite{ahmed1974discrete, strang1999discrete}. The type-II DCT (Eq.~\ref{eq: DCT}) and its inverse (IDCT) (Eq.~\ref{eq: IDCT}) are commonly used as the general DCT and the inverse DCT. In detail, the DCT of  a sequence $\mathbf{x}=[\mathbf{x}[0]\ \mathbf{x}[1]\ ...\ \mathbf{x}[N-1]]^T$ is computed as
	\begin{equation}
		\mathbf{X}[k] = \sum_{n=0}^{N-1}\mathbf{x}[n]\text{cos}\left[\frac{\pi}{N}\left(n+\frac{1}{2}\right) k\right],\label{eq: DCT}
	\end{equation}
	and its inverse is computed as
	\begin{equation}
		\mathbf{x}[n] = \frac{1}{N}\mathbf{X}[0]+\frac{2}{N}\sum_{k=1}^{N-1}\mathbf{X}[k]\text{cos}\left[\frac{\pi}{N}\left(n+\frac{1}{2}\right)k\right],\label{eq: IDCT}
	\end{equation}
	for $0\leq k \leq N-1, 0\leq n \leq N-1.$ 
	
\begin{figure}[htbp]
		\centering
		\includegraphics[width=0.7\linewidth]{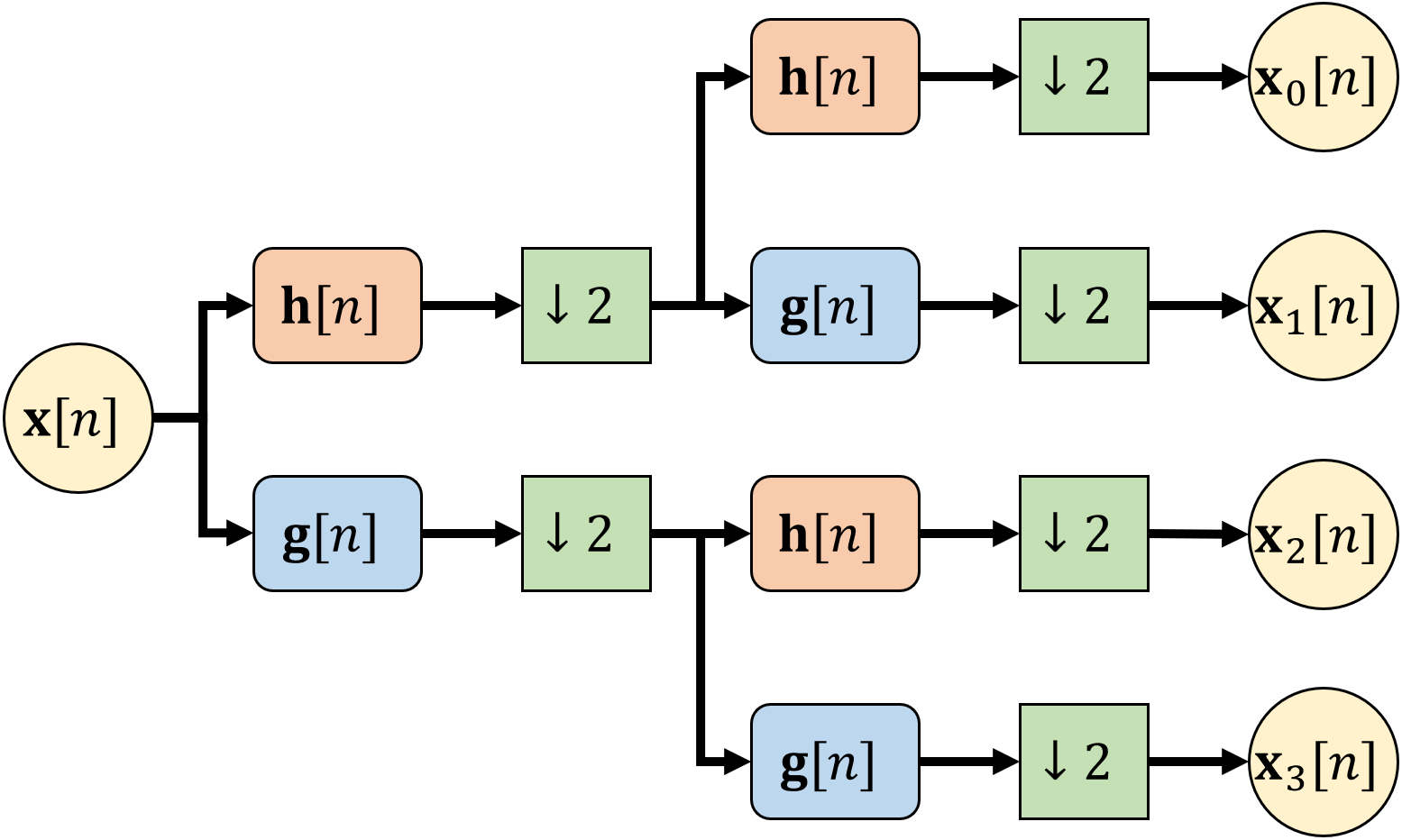}
		\caption{$N=4$ band ($M=2$ stage) subband decomposition filter bank structure for the orthogonal transform. The length of $\mathbf{x}$ is $N$. The length of each $\mathbf{x}_i$ is $\frac{N}{4}$ for $i=0, 1, 2, 3$. $\mathbf{X}=\{\mathbf{x}_0, \mathbf{x}_1, \mathbf{x}_2, \mathbf{x}_3\}$ is the $N$-length output of the orthogonal transform. In HT, $\mathbf{h}[n] = \{ 1, 1 \}$ and $\mathbf{g}[n] = \{ -1, 1 \}$. In BWT using Bior 1.3 coefficients, $\mathbf{h}[n] = \{-0.125, 0.125, 1, 1, 0.125, -0.125\}$ and $g[n] = \{-1, 1\}$~\cite{singh2011jpeg}.
		}
		\label{fig: filterbank}
	\end{figure}

	On the other hand, the Hadamard transform (HT) can be considered a binary version of the DFT. No multiplication is required if we ignore normalization. The HT is identical to the inverse HT (IHT) if the same normalization factor$\sqrt{\frac{1}{N}}$ is applied:
	\begin{equation}
		\mathbf{X}=\sqrt{\frac{1}{N}}\mathbf{H}_N\mathbf{x},\  \mathbf{x}=\sqrt{\frac{1}{N}}\mathbf{H}_N\mathbf{X},\label{eq: HT}
	\end{equation}
	where, $\mathbf{X}=[\mathbf{X}[0],\ ...,\ \mathbf{X}[N-1]]^T$. $\mathbf{H}_N$ is the $N$ by $N$ Hadamard matrix that can be constructed as
	\begin{equation}\label{eq: Hadamard matrix}
		\mathbf{H}_N = 
		\begin{cases}
			1,& N = 1,\\
			\begin{bmatrix}
				1 & 1 \\ 1 & -1
			\end{bmatrix},& N =2,\\
			\begin{bmatrix}
				\mathbf{H}_{\frac{N}{2}} & \mathbf{H}_{\frac{N}{2}} \\ \mathbf{H}_{\frac{N}{2}} & -\mathbf{H}_{\frac{N}{2}}
			\end{bmatrix}=\mathbf{H}_2 \otimes\mathbf{H}_{\frac{N}{2}},& N \ge 4,
		\end{cases}
	\end{equation}
	where, $\otimes$ stands for the Kronecker product.
	In the implementation, we can combine two $\sqrt\frac{1}{N}$ normalization terms from the HT and its inverse to one $\frac{1}{N}$ to avoid the square-root operation. 
	The Hadamard transform $\mathbf{H}_4$
	can be implemented using a wavelet filter bank with binary convolutional filters $\mathbf{h}[n] = \{ 1, 1 \}$ and the high pass filter $\mathbf{g}[n] = \{-1, 1 \}$ and dropout in two stages \cite{cetin1993block} as shown in Fig.~\ref{fig: filterbank}. This process is the basis of the fast algorithm. An $N=2^M$
	can be implemented in $M$ stages. In fact the low-pass filter
	$\mathbf{h}[n] = \{ 1, 1 \}$ and the high pass filter $\mathbf{g}[n] = \{-1, 1 \}$ are the building blocks of the butterfly operation of the fast Fourier transform (FFT) algorithm and the first stage of the Hadamard and the FFT are the same. In the FFT algorithm, complex exponential terms are also used in addition to butterflies in other stages. Each stage of the convolutional filter bank shown in Fig.~\ref{fig: filterbank} can be also expressed as a matrix-vector multiplication as in the FFT algorithm. 
	
	We can also use the wavelet filter bank to implement orthogonal and biorthogonal transforms such as ``Bior 1.3'', whose coefficients $\mathbf{h}[n] = \{-0.125, 0.125, 1, 1, 0.125, -0.125\}$ and $g[n] = \{-1, 1\}$~\cite{singh2011jpeg}. 

	The HT and DCT can be implemented using butterfly operations described in Eq. (1) in~\cite{fino1976unified} and the filter bank implementation of BWT is also fast \cite{cetin1993block}. Therefore, the complexity of each transform on a $N$-length vector is $O(N\log_2 N)$. The 2-dimensional (2D) transform is obtained from the 1-dimensional (1D) transform in a separable manner for computational efficiency. Thus, the complexity of a 2D transform on an $N\times N$ image is $O(N^2\log_2 N)$~\cite{vetterli1985fast}. Furthermore, using the hybrid quantum-classical approach~\cite{shukla2022hybrid}, the 1D-HT and the 2D-HT can be implemented in $O(N)$ and $O(N^2)$ time, respectively, as a quantum logic gate can be implemented efficiently in parallel. 
	
	
	The DCT convolution theorem~\ref{dct convolution theorem} is related to the DFT convolution theorem:
	\begin{theorem}[DCT convolution theorem]\label{dct convolution theorem}
		Let $N\in\mathbb{N}_+$, and $\mathbf{a}, \mathbf{x}\in\mathbb{R}^N$. Then, $\mathbf{y}=\mathbf{a}*_s\mathbf{x} \Longleftrightarrow \mathbf{Y}[k] = \mathbf{A}[k]\mathbf{X}[k]$, where, $\mathbf{Y}=\mathcal{D}(\mathbf{y}), \mathbf{A}=\mathcal{D}(\mathbf{a}), \mathbf{X}=\mathcal{D}(\mathbf{x}).$ 
	\end{theorem}
	Where, $\mathcal{D}(\cdot)$ stands for DCT, $*_s$ stands for the symmetric convolution.
	The relation between the symmetric convolution $*_s$ and the linear convolution $*$ is given by
	\begin{equation}
		\mathbf{a}*_s\mathbf{x} = \mathbf{\tilde{a}}*\mathbf{x}
	\end{equation}
	where $\mathbf{\tilde{a}}$ is the symmetrically extended kernel in $\mathbb{R}^{2N-1}$:
	\begin{equation}
		\mathbf{\tilde{a}}_k = \mathbf{a}_{|N-1-k|},
	\end{equation}
	for $k=0, 1, ..., 2N-2$. The proof of Theorem~\ref{dct convolution theorem} was presented in Section III in~\cite{shen1998dct}. 
	
	Similarly, we have the HT convolution theorem~\ref{ht convolution theorem}:
	\begin{theorem}[HT convolution theorem]\label{ht convolution theorem}
		Let $M\in\mathbb{N}$, $N=2^M$, and $\mathbf{a}, \mathbf{x}\in\mathbb{R}^N$. Then $ \mathbf{y}=\mathbf{a}*_d\mathbf{x} \Longleftrightarrow \mathbf{Y}[k] = \mathbf{A}[k]\mathbf{X}[k]$, where, $\mathbf{Y}=\mathcal{H}(\mathbf{y})$, $\mathbf{A}=\mathcal{H}(\mathbf{a})$, $\mathbf{X}=\mathcal{H}(\mathbf{x})$.
	\end{theorem}
	Where, $\mathcal{H}(\cdot)$ stands for HT, $*_d$ represents the dyadic convolution. Although the dyadic convolution $*_d$ is not the same as circular convolution, we can still use the HT for convolutional filtering in neural networks. This is because 
	HT is also related to the block Haar wavelet packet transform~\cite{cetin1993block} and each Hadamard coefficient approximately represents a frequency band. As a result, applying weights onto frequency bands and computing the IHT is an approximate way of frequency domain filtering similar to the Fourier transform-based convolutional filtering. The proof of Theorem~\ref{ht convolution theorem} is presented in~\cite{uvsakova2002walsh}.
	
	In brief, convolution theorems have the property that convolution in the time or space domain is equivalent to element-wise multiplication in the transform domain. This property inspires us to design the neural network layers to replace the Conv2D layers in a deep neural network.

	
	\subsection{Transform-Based Perceptron Layers}
\begin{figure}[t]
		\centering
		\subfloat[\label{fig: single-pod}1-channel.]{\includegraphics[width=0.243\linewidth]{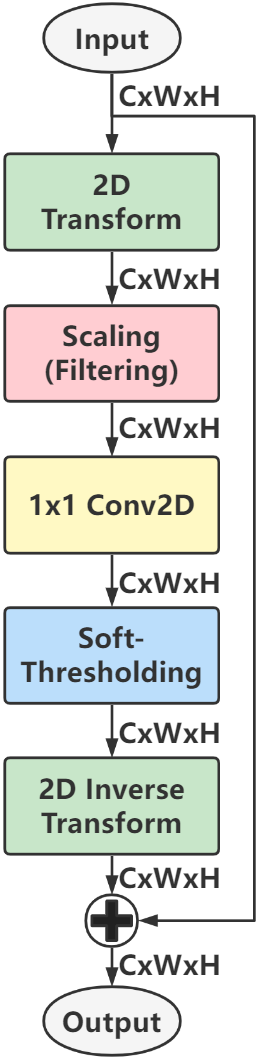}}\hspace{1pt}
		\subfloat[\label{fig: 3-pod}Multi-channel.]{\includegraphics[width=0.545\linewidth]{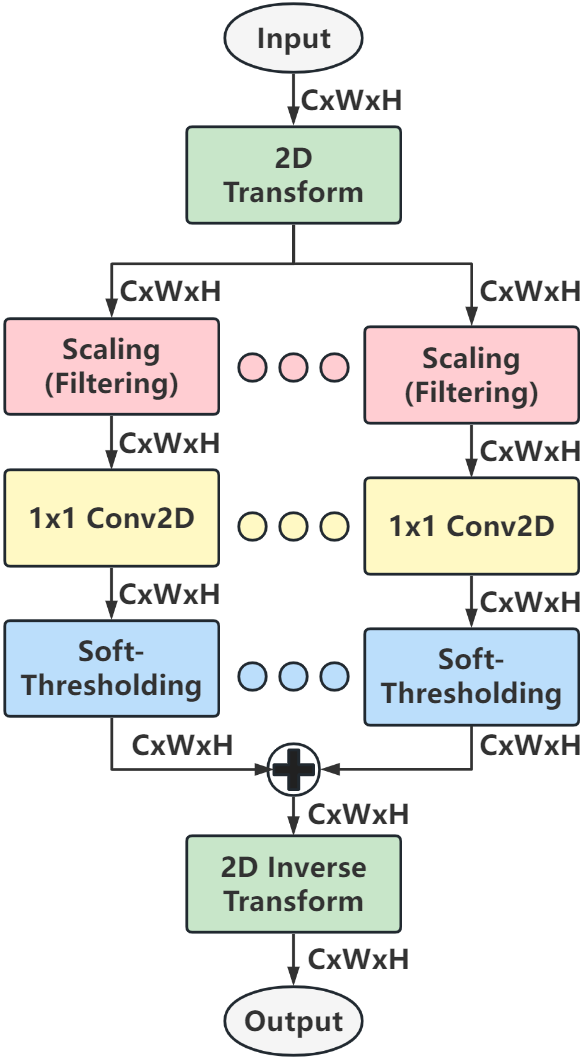}}
		\caption{(a) Single-channel perceptron layer and (b) multi-channel perceptron layer. Zero padding and truncation are applied in the HT-perceptron layers and the BWT-perceptron layers if the image size is not a power of 2. 
		}
		\label{fig: perceptron}
	\end{figure}

	Fig.~\ref{fig: perceptron} presents the structures of the proposed transform-based perceptron layers. We propose two types of perceptron layers for each transform. One is a single-channel layer as shown in Fig.~\ref{fig: single-pod}, and the other is a multi-channel layer as shown in Fig.~\ref{fig: 3-pod}. The multi-channel layer contains multiple times parameters as the single-channel layer, but it can produce a higher accuracy according to our experiments. In the transform domain of each channel, a scaling layer performs the filtering operation by element-wise multiplication. Then a $1\times 1$ Conv2D layer performs channel-wise filtering.
	After this step, there is a trainable soft-thresholding layer denoising the data and acting as the nonlinearity. In the single-channel layer, a shortcut connection is implemented. In the multi-channel layer, the outputs of all channels are summed up. We do not implement the shortcut connection in the multi-channel layer because there are multiple channels replacing the shortcut and they effectively transmit the derivatives to earlier layers. The 2D transform and its inverse along the width and height convert the tensor between the space domain and the transform domain. 
	
	
	\begin{figure}[t]
		\centering
		\includegraphics[width=1\linewidth]{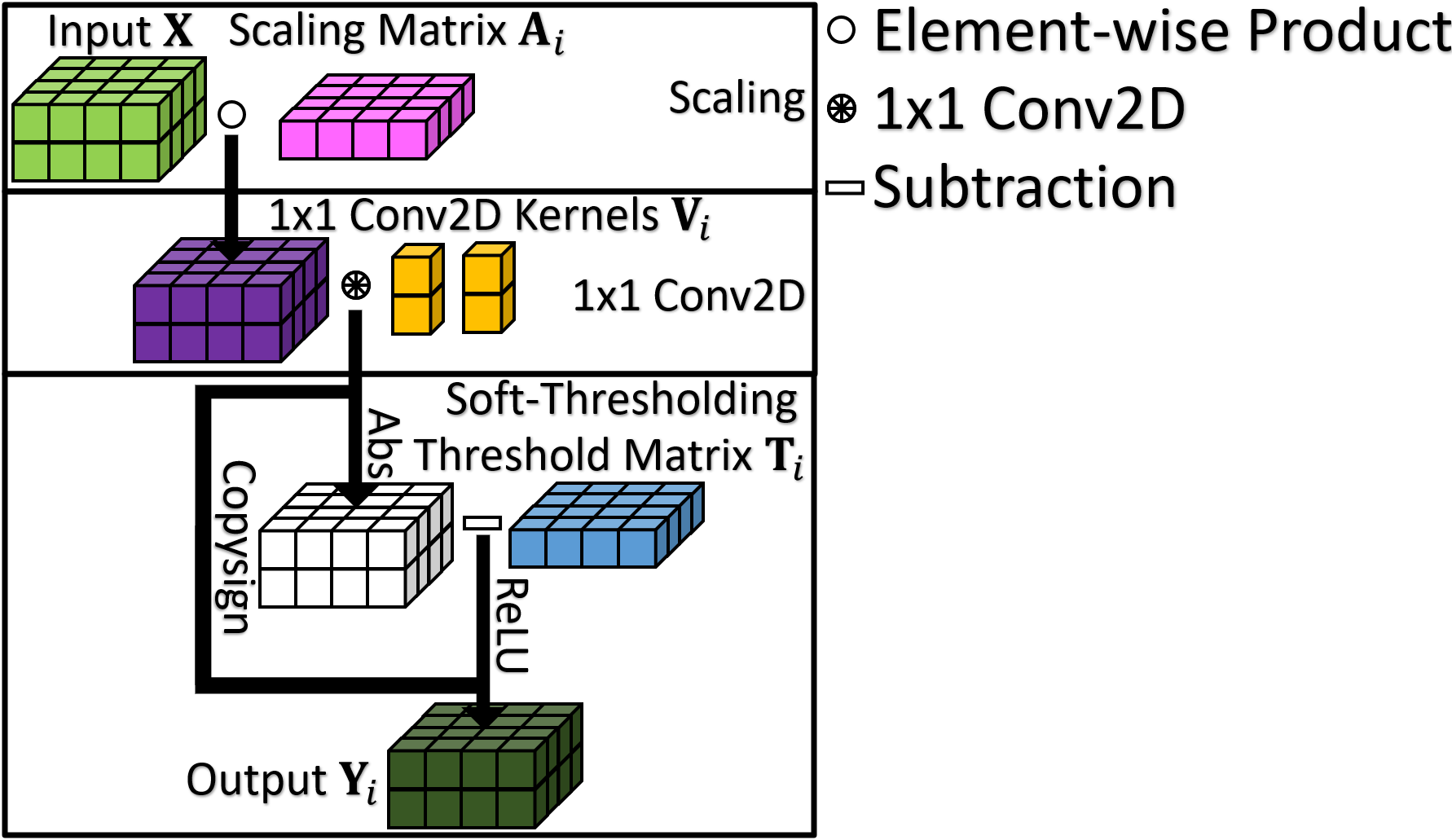}
		\caption{Processing in the transform domain. }
		\label{fig: pod}
	\end{figure}
	
	The scaling layer is derived from Theorems~\ref{dct convolution theorem} and~\ref{ht convolution theorem}, which state that the convolution in the space domain is equivalent to the element-wise multiplication in the transform domain. In detail, the input tensor in $\mathbb{R}^{C\times W \times H}$ is element-wise multiplied with a weight matrix in $\mathbb{R}^{W \times H}$ to perform an operation equivalent to the spatial domain convolutional filtering. 
	
	The trainable soft-thresholding layer is applied to remove small entries in the transform domain. It is similar to image coding and transform-domain denoising. It is defined as:
	\begin{equation}
		\mathbf{y} = \mathcal{S}_T(\mathbf{x}) = \text{sign}(\mathbf{x})(|\mathbf{x}|-T)_+,
	\end{equation}
	where, $T$ is a non-negative trainable threshold parameter, $(\cdot)_+$ stands for the ReLU function. An input tensor in $\mathbb{R}^{C\times W \times H}$ is computed with threshold parameters in $\mathbb{R}^{W \times H}$, and these threshold parameters are determined using the back-propagation algorithm. The ReLU and its variants such as leaky ReLU, Sigmoid Linear Unit (SiLU)~\cite{hendrycks2016gaussian, javid2022developing} functions are not suitable because both positive and negative values are important in the transform domain. A completely positive waveform can have both significant positive and negative values in the transform domain. Therefore, we should use an antisymmetric function, \textit{i.e.}, $f(x) = - f(-x)$, with respect to 0 in our networks. Our ablation study on the CIFAR-10 dataset experimentally shows that the soft-thresholding function is superior to the ReLU, the Leaky ReLU, and the SiLU functions in the transform-based layers.
	
	The overall process of each channel for the transform domain analysis (scaling, $1\times1$ Conv2D, and soft-thresholding) is depicted in Figure~\ref{fig: pod}. The single-channel and the $P$-channel transform-based perceptron layers are computed as:
	\begin{equation}    
		\mathbf{y}=\mathcal{T}^{-1}\left(\mathcal{S}_\mathbf{T}\left(\mathcal{T}(\mathbf{x})\circ\mathbf{A}\circledast\mathbf{V}\right)\right)+\mathbf{x},
	\end{equation}
	\begin{equation}    
		\mathbf{y}=\mathcal{T}^{-1}\left(\sum_{i=0}^{P-1}\mathcal{S}_{\mathbf{T}_i}\left(\mathcal{T}(\mathbf{x})\circ\mathbf{A}_i\circledast\mathbf{V}_i\right)\right),
	\end{equation}
	where, $\mathcal{T}(\cdot)$ and $\mathcal{T}^{-1}(\cdot)$ stand for 2D transform (DCT, HT, or BWT) and 2D inverse transform, $\mathbf{A}$ and $\mathbf{A}_i$ are the scaling matrices, $\mathbf{V}$ and $\mathbf{V}_i$ are the kernels in the $1\times1$ Conv2D layers, $\mathbf{T}$ and $\mathbf{T}_i$ are the threshold parameter matrices in the soft-thresholding, $\circ$ stands for the element-wise multiplication, and $\circledast$ represents the $1\times 1$ 2D convolution over an input composed of several input planes performed using PyTorch's Conv2D API.

	\begin{table*}[b]
		\caption{Parameters and Multiply–Accumulate (MACs) for a $C$-channel $N\times N$ tensor.}
		\centering
		\begin{tabular}{lcc}
			\hline\noalign{\smallskip}
			\bf{Layer (Operation)}&\bf{Parameters}&\bf{MACs}\\
			\noalign{\smallskip}\hline\noalign{\smallskip}
			$K\times K$ Conv2D& $K^2C^2$&$K^2N^2C^2$\\
			$3\times 3$ Conv2D& $9C^2$&$9N^2C^2$\\
			\noalign{\smallskip}\hline\noalign{\smallskip}
			2D-DCT/BWT/IDCT/IBWT (fast algorithm~\cite{cetin1993block,vetterli1985fast})&$0$& $\left(\frac{5N^2}{2}\log_2 N+\frac{N^2}{3}-6N+\frac{62}{3}\right)C$\\
			2D-DCT/BWT/IDCT/IBWT (matrix-vector product, used in this work.)&$0$& $2N^3C$\\
			Scaling, Soft-Thresholding& $2N^2$&$N^2C$\\
			$1\times 1$ Conv2D & $C^2$ & $N^2C^2$\\
			\bf{$P$-Channel DCT/BWT-Perceptron}& $2PN^2+PC^2$&$4N^3C+PN^2C^2+PN^2C$\\  
			\bf{$P$-Channel HT-Perceptron}& $2PN^2+PC^2$&$PN^2C^2+PN^2C$\\      
			\noalign{\smallskip}\hline\noalign{\smallskip}
			\multicolumn{3}{p{0.66\textwidth}}{Note: MACs from 2D-HT are omitted because there is no multiplication.} 
		\end{tabular}
		\label{tab: parameters and MACs}
	\end{table*}
	
	\subsection{Comparison with the Convolutional Layer}
	\subsubsection{Spatial and Channel Characteristics Comparison}
	The Conv2D layer has spatial-agnostic and channel-specific characteristics. Because of the spatial-agnostic characteristics of a Conv2D layer, the network cannot adapt different visual patterns corresponding to different spatial locations. On the contrary, the transform-based perceptron layer is location-specific and channel-specific. The 2D transform is location-specific but channel-agnostic, as it is computed using the entire block as a weighted summation on the spatial feature map. The scaling layer is also location-specific but channel-agnostic, as different scaling parameters (filters) are applied on different entries, and weights are shared in different channels. That is why we also use PyTorch's $1\times 1$ Conv2D to make the transform-based perceptron layer channel-specific.
	
	Because the Conv2D layer is spatial-agnostic, the convolutional kernels cannot utilize the information from the location of the pixels. Conversely, the transform-based perceptron layer is location-specific, which means each weight parameter gains the location information.


	\subsubsection{Number of Parameters and MACs Comparison}
	The comparison of the number of parameters and  Multiply–Accumulate (MACs) are presented in Table~\ref{tab: parameters and MACs}. In a Conv2D layer, we need to compute $N^2K^2$ multiplications for a $C$-channel feature map of size $N\times N$ with a kernel size of $K\times K$. In the DCT-perceptron layer, there are $N^2$ multiplications from the scaling and $N^2$ multiplications from the $1\times1$ Conv2D layer. There is no multiplication in the soft-thresholding function because the product between the sign of the input and the subtraction result can be implemented using the copysign operation. 
	Although the 2D-DCT and the 2D-IDCT can be implemented in $O(N^2C\log_2N)$ using the fast algorithm~\cite{vetterli1985fast}, this fast approach is not officially supported in PyTorch. In this paper, we implement the 2D-DCT and the 2D-IDCT using the product between the DCT/IDCT weight matrix and the input tensor. Its complexity is $O(N^3C)$, and its MACs are $2N^3C$. Therefore, A $P$-channel DCT-perceptron layer has $2PN^2+PC^2$ parameters with $4N^2C+PN^2C^2+PN^2C$ MACs. 
	
	The BWT-perceptron layer has the same number of parameters and MACs as the DCT-perceptron layer. As for the HT-perceptron layers, because the HT is a multiple-free transform, there are only $2PN^2+PC^2$ parameters with $PN^2C+PN^2C^2$ MACs in the $P$-channel HT-perceptron layer. A 3-channel perceptron layer has the same amount of parameters but fewer MACs compared to a $3\times 3$ Conv2D layer if $C=N$. Therefore, in the experiment section, we mainly use the 3-channel perception layer to compare with the Conv2D layer. Furthermore, in most main-stream CNNs, $C$ is usually much larger than $N$ in the hidden layers, then our proposed perceptron layers can reduce parameters and computational cost for these CNNs even with the matrix-vector product for the transforms.
	
	\subsection{Introduce Transform-based Perceptron Layers into ResNets}
\begin{figure}[t]
		\centering
		\subfloat[\label{fig: ResNet Block}ResNet Conv2D blocks.]{\includegraphics[width=0.47\linewidth]{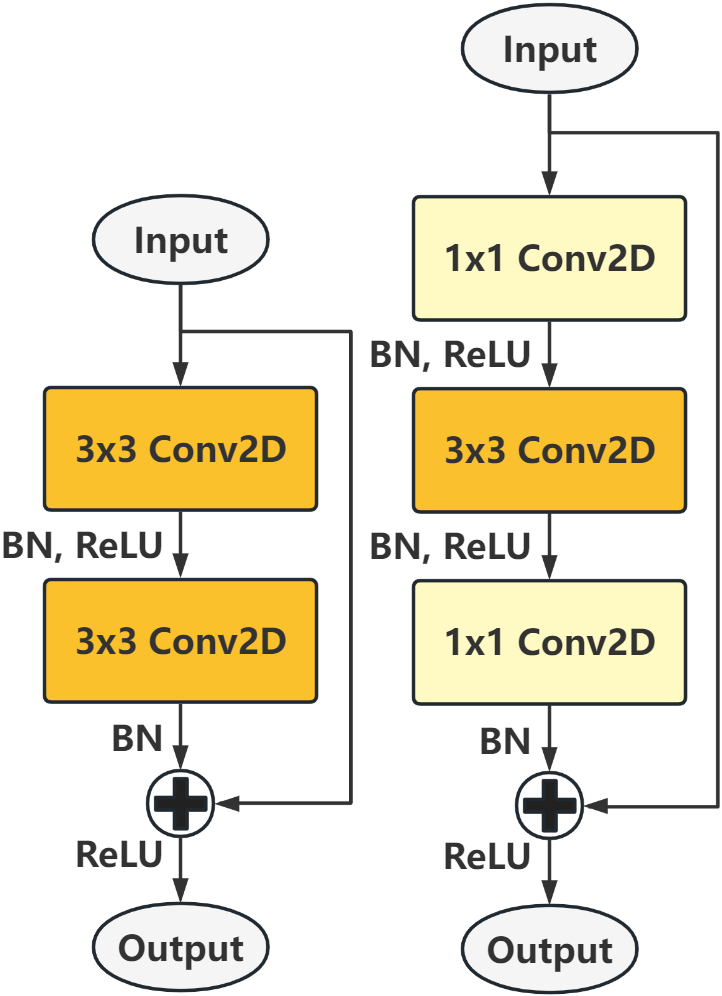}} \hspace{1pt}
		\subfloat[\label{fig: ResNet-P Block}ResNet TP blocks.]{\includegraphics[width=0.47\linewidth]{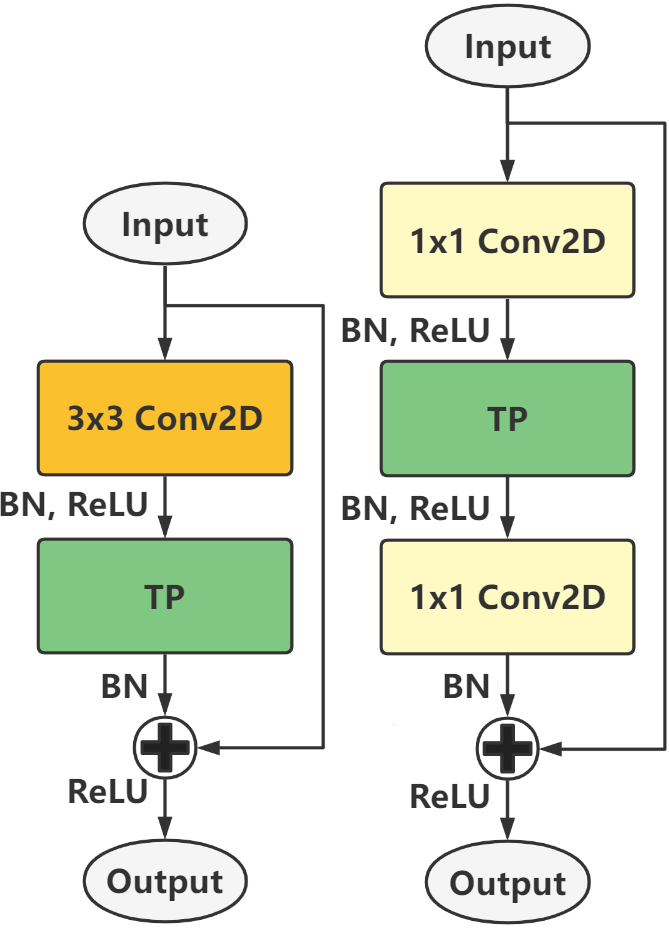}}
		\caption{(a) ResNet's convolutional residual blocks V1 (left) and V2 (right) versus (b) the proposed transform-based Perceptron (TP) residual blocks V1 (left) and V2 (right). BN stands for batch normalization. One $3\times3$ Conv2D layer in each block is replaced by one proposed DCT-perceptron layer. }
		\label{fig: block}
	\end{figure}

	In this paper, we introduce the proposed layers into ResNet-18, ResNet-20, ResNet-50, and ResNet-101. 
	Fig.~\ref{fig: ResNet Block} shows the commonly used convolutional blocks in ResNets. The V1 version that contains two $3\times 3$ Conv2D layers is used in ResNet-18 and ResNet-20, and the V2 version that contains one $3\times 3$ and two one $1\times 1$ Conv2D layers is used in ResNet-50 and ResNet-101. As Fig.~\ref{fig: ResNet-P Block} shows, we replace the second $3\times 3$ Conv2D layer in each convolutional block V1, and the $3\times 3$ Conv2D layer in some convolutional blocks V2.
	
	\begin{table}[htbp]
		\caption{Structure of ResNet-20 with the transform-based perceptron (TP) layer for CIFAR-10 classification. }
		\centering
		\begin{tabular}{lcc}
			\hline\noalign{\smallskip}
			\bf{Layer}&\bf{Output Shape}&\bf{Implementation Details}\\
			\noalign{\smallskip}\hline\noalign{\smallskip}
			Input&$3\times32\times32$&-\\
			Conv1&$16\times32\times32$&$3\times3, 16$\\
			\bf{Conv2\_x}&$16\times32\times32$&$\left[ \begin{array}{c} 3\times3, 16  \\ \bf{\text{TP}, 16} \end{array}\right]\times 3$\\
			\bf{Conv3\_x}&$32\times16\times16$&$\left[ \begin{array}{c} 3\times3, 32  \\ \bf{\text{TP}, 32} \end{array}\right]\times 3$\\
			\bf{Conv4\_x}&$64\times8\times8$&$\left[ \begin{array}{c} 3\times3, 32  \\ \bf{\text{TP}, 64} \end{array}\right]\times 3$\\
			GAP&$64$&Global Average Pooling\\
			Output&$10$&Linear\\
			\noalign{\smallskip}\hline\noalign{\smallskip}
			\multicolumn{3}{p{0.95\linewidth}}{Note: Building blocks are shown in brackets, with the numbers of blocks stacked. Downsampling with a stride of 2 is performed by Conv3\_1 and Conv4\_1.}
		\end{tabular}
		\label{tab: resnet-20}
	\end{table}
	
	\begin{table}[t]
		\caption{Structure of ResNet-18 with the transform-based perceptron (TP) layers for ImageNet-1K classification.}
		\centering
		\begin{tabular}{lcc}
			\hline\noalign{\smallskip}
			\bf{Layer}&\bf{Output Shape}&\bf{Implementation Details}\\
			\noalign{\smallskip}\hline\noalign{\smallskip}
			Input&$3\times224\times224$&-\\
			Conv1&$64\times112\times112$&$7\times7, 64$, stride 2\\
			MaxPool&$64\times56\times56$&$2\times2$, stride 2\\
			\bf{Conv2\_x}&$64\times56\times56$&$\left[ \begin{array}{c} 3\times3, 64  \\ \bf{\text{TP}, 64} \end{array}\right]\times 2$\\
			\bf{Conv3\_x}&$128\times28\times28$&$\left[ \begin{array}{c} 3\times3, 128  \\ \bf{\text{TP}, 128} \end{array}\right]\times 2$\\
			\bf{Conv4\_x}&$256\times14\times14$&$\left[ \begin{array}{c} 3\times3, 256  \\ \bf{\text{TP}, 256} \end{array}\right]\times 2$\\
			\bf{Conv5\_x}&$512\times7\times7$&$\left[ \begin{array}{c} 3\times3, 512  \\ \bf{\text{TP}, 512} \end{array}\right]\times 2$\\
			GAP&$512$&Global Average Pooling\\
			Output&$1000$&Linear\\
			\noalign{\smallskip}\hline\noalign{\smallskip}
			
			\multicolumn{3}{p{0.95\linewidth}}{Note:  Building blocks are shown in brackets, with the numbers of blocks stacked. Downsampling with a stride of 2 is performed by Conv3\_1, Conv4\_1, and Conv5\_1. }
		\end{tabular}
		\label{tab: resnet-18}
	\end{table}

\begin{table}[t]
	\caption{Structure of ResNet-50 with the transform-based perceptron (TP) layers for ImageNet-1K classification. }
	\centering   
	\begin{tabular}{lcc}
		\hline\noalign{\smallskip}
		\bf{Layer}&\bf{Output Shape}&\bf{Implementation Details}\\
		\noalign{\smallskip}\hline\noalign{\smallskip}
		Input&$3\times224\times224$&-\\
		Conv1&$64\times112\times112$&$7\times7, 64$, stride 2\\
		MaxPool&$64\times56\times56$&$2\times2$, stride 2\\
		Conv2\_1&$256\times56\times56$&$\left[ \begin{array}{c} 1\times1, 64  \\ 3\times3, 64, \\ 1\times1, 256 \end{array}\right]$\\
		\bf{Conv2\_2}&$256\times56\times56$&$\left[ \begin{array}{c} 1\times1, 64  \\ \bf{\text{TP}, 64} \\ 1\times1, 256 \end{array}\right]$\\
		Conv2\_3&$256\times56\times56$&$\left[ \begin{array}{c} 1\times1, 64  \\ 3\times3, 64 \\ 1\times1, 256 \end{array}\right]$\\
		Conv3\_1&$512\times28\times28$&$\left[ \begin{array}{c} 1\times1, 128  \\ 3\times3, 128, \text{stride }2 \\ 1\times1, 512 \end{array}\right]$\\
		\bf{Conv3\_2}&$512\times28\times28$&$\left[ \begin{array}{c} 1\times1, 128  \\ \bf{\text{TP}, 128} \\ 1\times1, 512 \end{array}\right]$\\
		Conv3\_3&$512\times28\times28$&$\left[ \begin{array}{c} 1\times1, 128  \\ 3\times 3, 128 \\ 1\times1, 512 \end{array}\right]$\\
		\bf{Conv3\_4}&$512\times28\times28$&$\left[ \begin{array}{c} 1\times1, 128  \\ \bf{\text{TP}, 128} \\ 1\times1, 512 \end{array}\right]$\\
		Conv4\_1&$1024\times14\times14$&$\left[ \begin{array}{c} 1\times1, 256  \\ 3\times3, 256, \text{stride }2 \\ 1\times1, 1024 \end{array}\right]$\\
		\bf{Conv4\_2}&$1024\times14\times14$&$\left[ \begin{array}{c} 1\times1, 256  \\ \bf{\text{TP}, 256} \\ 1\times1, 1024 \end{array}\right]$\\
		Conv4\_3&$1024\times14\times14$&$\left[ \begin{array}{c} 1\times1, 256  \\ 3\times 3, 256 \\ 1\times1, 1024 \end{array}\right]$\\
		\bf{Conv4\_4}&$1024\times14\times14$&$\left[ \begin{array}{c} 1\times1, 256  \\ \bf{\text{TP}, 256} \\ 1\times1, 1024 \end{array}\right]$\\
		Conv4\_5&$1024\times14\times14$&$\left[ \begin{array}{c} 1\times1, 256  \\ 3\times 3, 256 \\ 1\times1, 1024 \end{array}\right]$\\
		\bf{Conv4\_6}&$1024\times14\times14$&$\left[ \begin{array}{c} 1\times1, 256  \\ \bf{\text{TP}, 256} \\ 1\times1, 1024 \end{array}\right]$\\
		Conv5\_1&$2048\times7\times7$&$\left[ \begin{array}{c} 1\times1, 512  \\ 3\times3, 512, \text{stride }2 \\ 1\times1, 2048 \end{array}\right]$\\		
		\bf{Conv5\_2}&$2048\times7\times7$&$\left[ \begin{array}{c} 1\times1, 512  \\ \bf{\text{TP}, 512} \\ 1\times1, 2048 \end{array}\right]$\\	
		Conv5\_3&$2048\times7\times7$&$\left[ \begin{array}{c} 1\times1, 512  \\ 3\times3, 512 \\ 1\times1, 2048 \end{array}\right]$\\	
		GAP&$2048$&Global Average Pooling\\
		Output&$1000$&Linear\\
		\noalign{\smallskip}\hline\noalign{\smallskip}
		\multicolumn{3}{p{0.98\linewidth}}{Note: In the CIFAR-100 task, the input is $3\times 32\times 32$. MaxPool is removed. Conv1 is implemented as ``$3\times3, 64$, stride 1". Output shapes of the layers are reduced correspondingly.}
	\end{tabular}
	
\label{tab: resnet-50 2}
\end{table}

To revise ResNets, we retain the first $3\times 3$ Conv2D layer, then we replace those $3\times 3$ Conv2D layers at the even indices (the second Conv2D in each convolutional block in ResNet-20 and ResNet-18, and Conv2\_2, Conv3\_2, Conv3\_4, etc., in ResNet-50 and ResNet-101) with the proposed transform-based perception layers. We keep the $3\times3$ Conv2D layers at odd indices because, in this way, we use the regular $3\times 3$ Conv2D layer and the proposed HT-perceptron layer by turns, then the network can extract features in different manners efficiently. Table~\ref{tab: resnet-20} describes the method we revising the ResNet-20 for the CIFAR-10 classification task, and Tables~\ref{tab: resnet-18} and~\ref{tab: resnet-50 2} present the revised ResNet-18 and ResNet-50 for the ImageNet-1K classification task.


In Sections~\ref{sec: Experimental Results}--\ref{sec: Conclusion}, to call the revised ResNets, we add $P$C-Transform in front of their name. For example, 3C-DCT-ResNet-20 is the ResNet-20 revised by the three-channel DCT-perceptron layer. The BWT-ResNets are implemented using Bior 1.3 coefficients~\cite{singh2011jpeg}.


\section{Experimental Results}\label{sec: Experimental Results}
\begin{figure}[htbp]
\centering
\subfloat[CIFAR-10\label{fig: CIFAR-10}]{\includegraphics[width=0.5\linewidth]{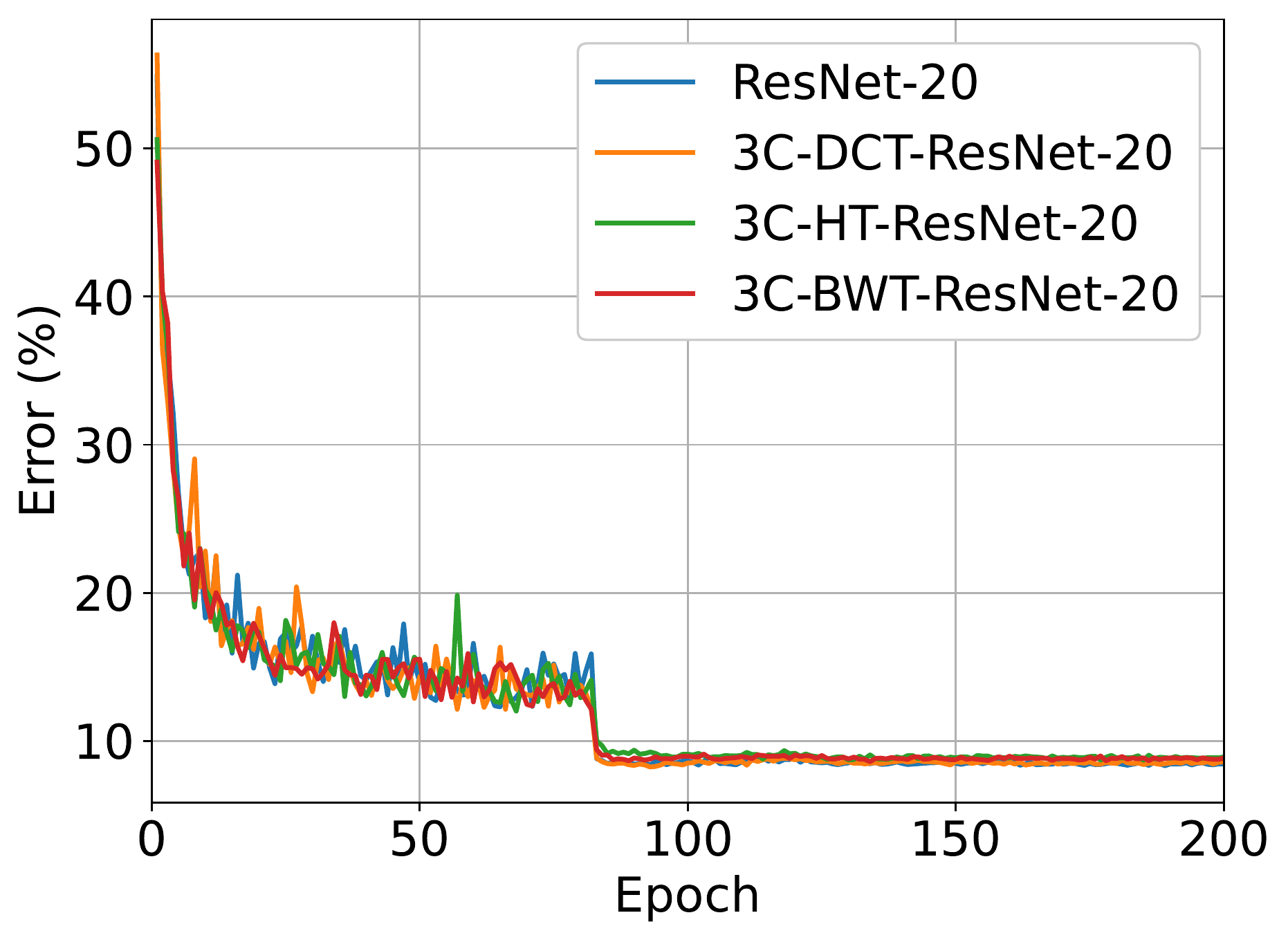}}
\subfloat[CIFAR-100\label{fig: CIFAR-100}]{\includegraphics[width=0.5\linewidth]{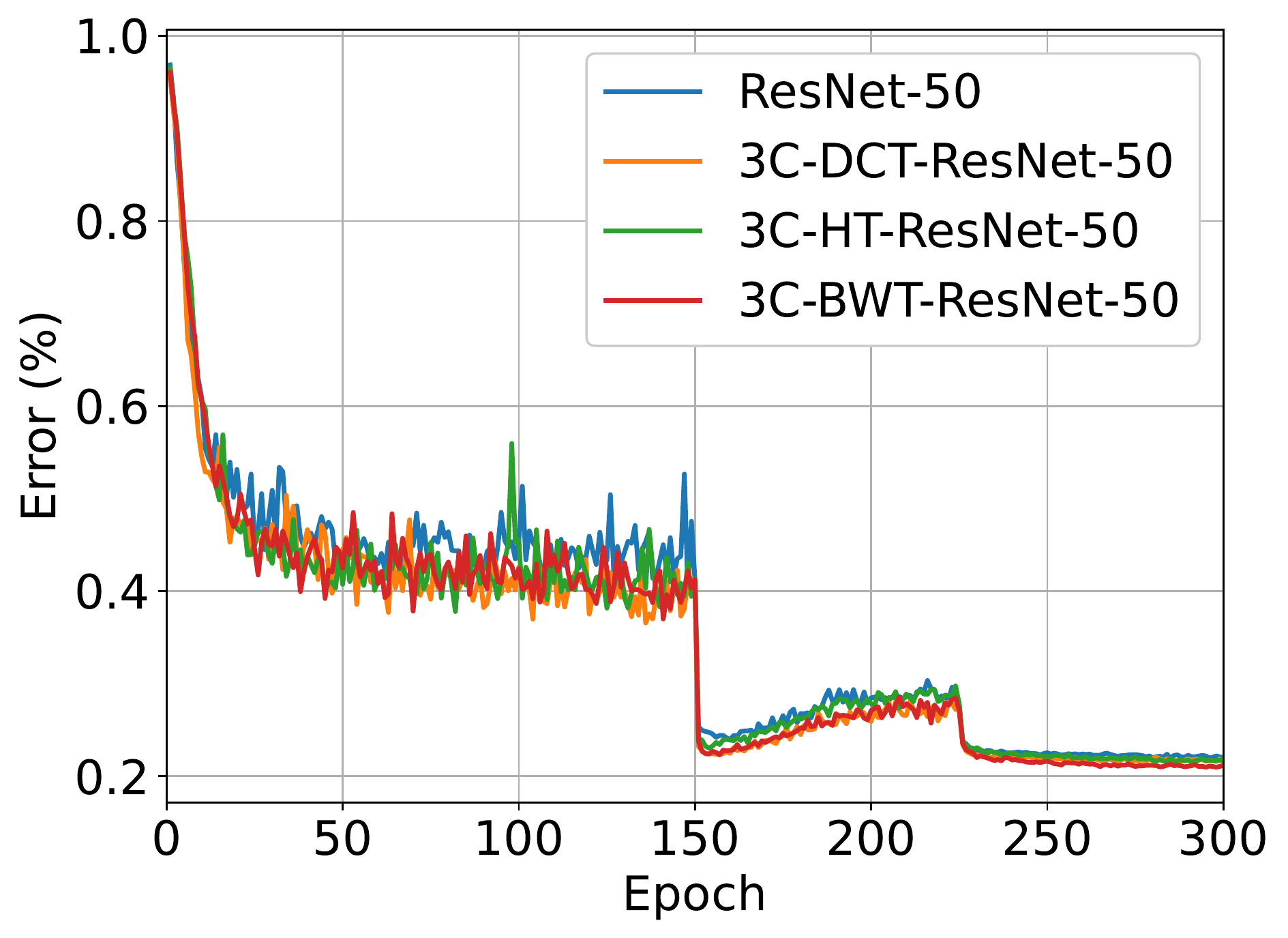}}\\
\subfloat[ImageNet-1K (224$^2$)\label{fig: ImageNet-1K}]{\includegraphics[width=0.5\linewidth]{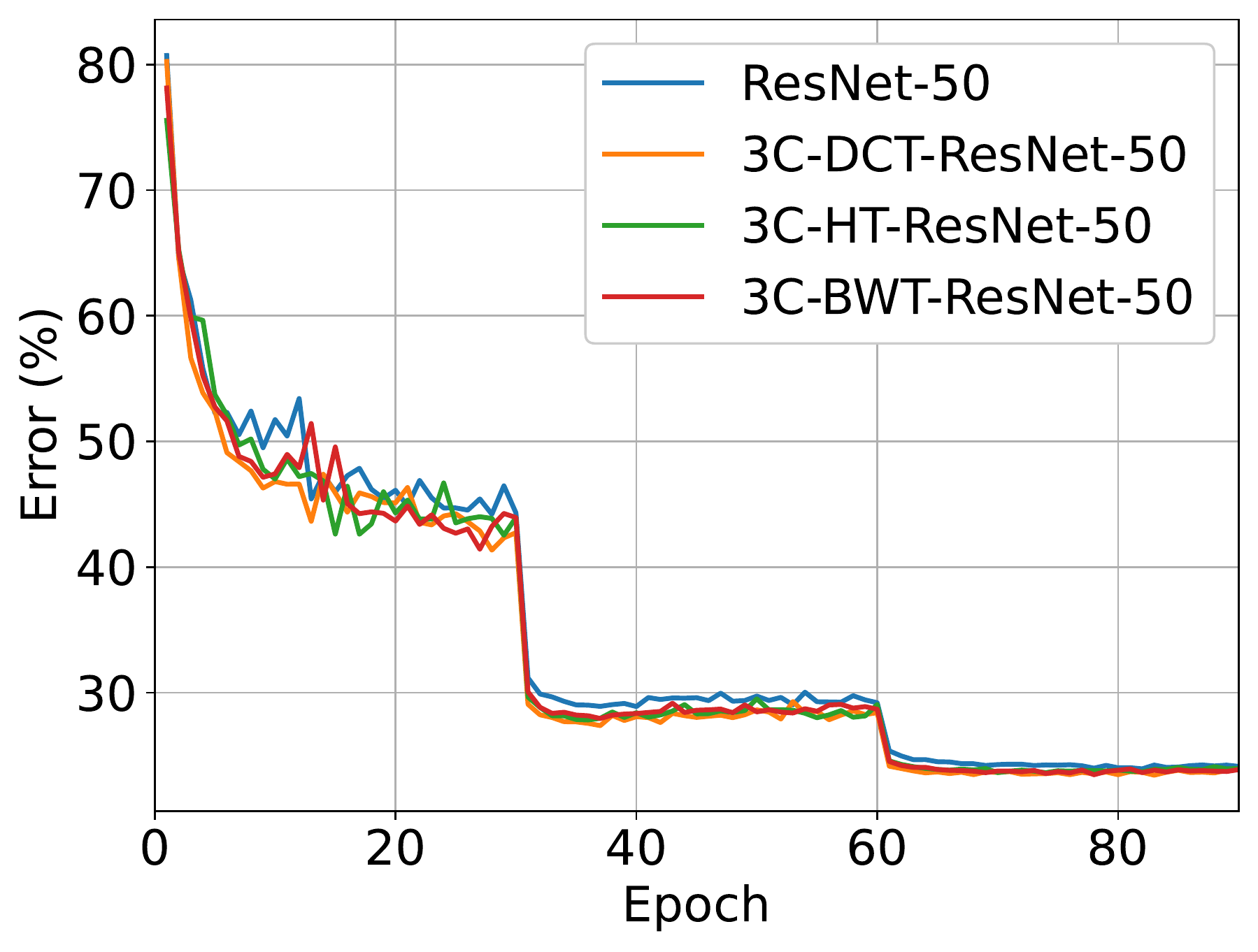}}
\subfloat[ImageNet-1K (224$^2$)\label{fig: ImageNet-1K_resnet101}]{\includegraphics[width=0.5\linewidth]{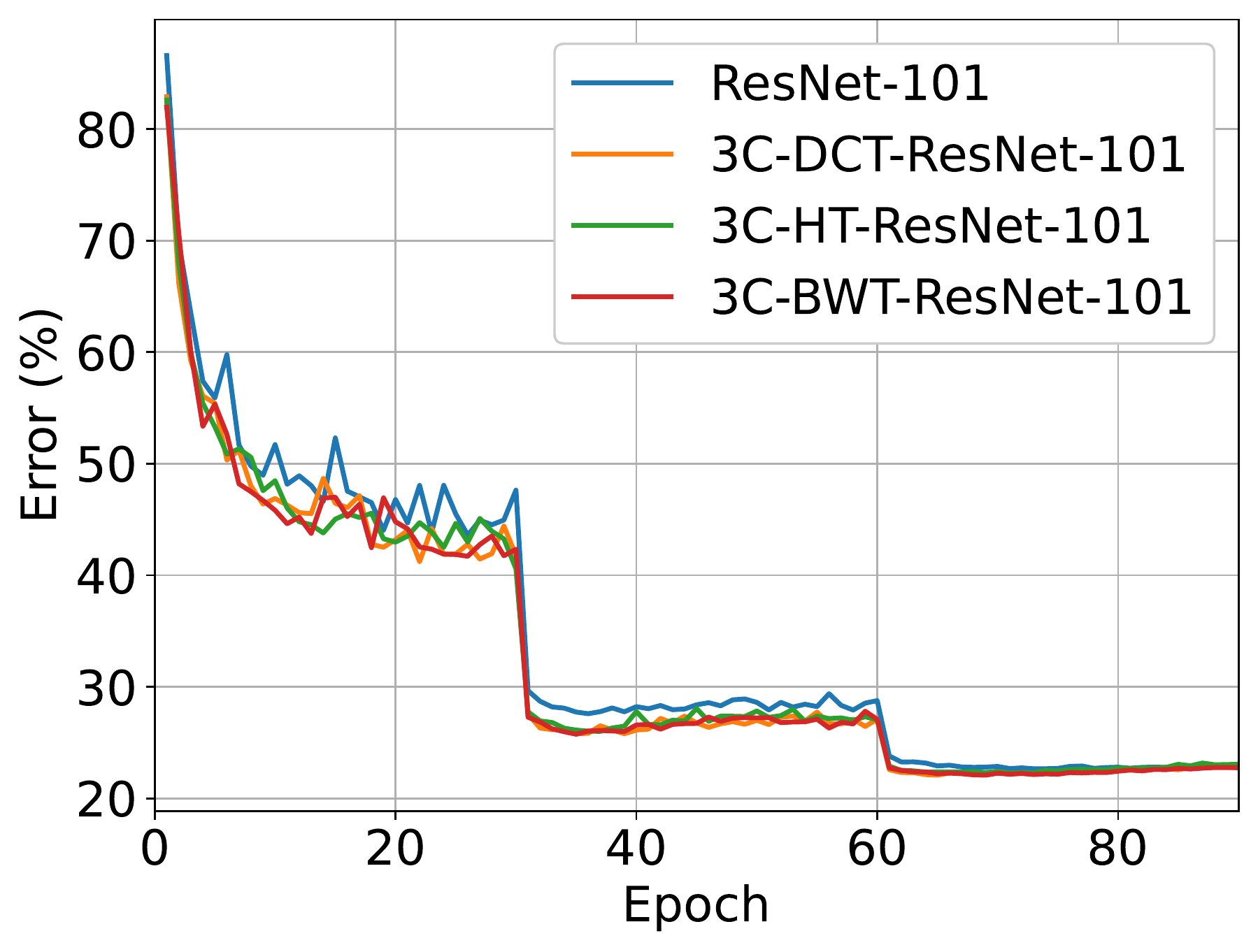}}
\caption{Training on CIFAR-10, CIFAR-100, and ImageNet-1K. In (a) and (b), curves denote test error. In (c) and (d), curves denote the validation error of the center crops. With reduced parameters and MACs, comparable CIFAR-10 accuracy results, better CIFAR-100 accuracy results, and better ImageNet-1K accuracy results are obtained using the proposed tri-channel DCT/HT/BWT-perceptron layers.}
\label{fig: training log}
\end{figure}

In this work, our ResNet-101 experiments are carried out on a server computer with an NVIDIA A100 GPU, and our other experiments are carried out on a workstation computer with an NVIDIA RTX 3090 GPU. The code is written in PyTorch in Python 3. The experiments include the CIFAR-10, CIFAR-100, and the ImageNet-1K classification task. In the CIFAR-10 experiments, ResNet-20 is used as the backbone network.  In the CIFAR-100 experiments, ResNet-50 is used as the backbone network. In the ImageNet-1K experiments, ResNet-18, ResNet-50, and ResNet-101 are used as the backbone networks. The CIFAR-10 and the CIFAR-100 experiments carry out our ablation study. Fig.~\ref{fig: training log} presents the test error history of our models.

\subsection{CIFAR-10 Experiments}
Training ResNet-20 and its revisions follow the implementation in~\cite{he2016deep}. In detail, we use an SGD optimizer with a weight decay of 0.0001 and momentum of 0.9. Models are trained with a mini-batch size of 128 for 200 epochs. The initial learning rate is 0.1, and the learning rate is reduced by a factor of 1/10 at epochs 82, 122, and 163, respectively. Data augmentation is implemented as follows: First, we pad 4 pixels on the training images. Then, we apply random cropping to get 32 by 32 images. Finally, we randomly flip images horizontally. We normalize the images with the means of [0.4914, 0.4822, 0.4465] and the standard variations of [0.2023, 0.1994, 02010]. During the training, the best models are saved based on the accuracy of the CIFAR-10 test dataset, and their accuracy numbers are reported in Table~\ref{tab: CIFAR-10}.

\begin{table*}[t]
\caption{CIFAR-10 Experimental Results.}
\centering
\begin{tabular}{lccc}
	\hline\noalign{\smallskip}
	\bf{Method}&\bf{Parameters}&\bf{MACs (M)}&\bf{Accuracy}\\
	\noalign{\smallskip}\hline\noalign{\smallskip}
	ResNet-20 \cite{he2016deep} (official)&0.27M&-&91.25\%\\
	HT-based ResNet-20~\cite{pan2022block}&133,082 (51.16\%$\downarrow$)&-&90.12\%\\
	\noalign{\smallskip}\hline\noalign{\smallskip}
	ResNet-20 (our trial, baseline)&272,474&41.32&91.66\%\\
	1C-DCT-ResNet-20&151,514 (44.39\%$\downarrow$)&30.79 (25.5\%$\downarrow$)&91.59\%\\
	3C-DCT-ResNet-20&199,898 (26.64\%$\downarrow$)&35.68 (13.6\%$\downarrow$)&\bf{91.75\%}\\
	1C-HT-ResNet-20&151,514 (44.39\%$\downarrow$)&22.53 (45.5\%$\downarrow$)&91.25\%\\
	3C-HT-ResNet-20&199,898 (26.64\%$\downarrow$)&27.42 (33.6\%$\downarrow$)&91.29\%\\
	3C-BWT-ResNet-20&199,898 (26.64\%$\downarrow$)&35.68 (13.6\%$\downarrow$)&91.35\%\\
	ResNet-20+1C-DCT-P.&276,826 (1.60\%$\uparrow$)&41.65 (0.8\%$\uparrow$)&91.82\%\\
	\noalign{\smallskip}\hline\noalign{\smallskip}
\end{tabular}
\label{tab: CIFAR-10}
\end{table*}

As Table~\ref{tab: CIFAR-10} shows, on the one hand, DCT-ResNet-20, which is implemented by revising Conv2D layers using the single-channel DCT-perceptron layers, contains 44.39\% fewer parameters than the baseline ResNet-20 model, but it only suffers a 0.07\% accuracy lost. On the other hand, the 3C-DCT-ResNet-20, which is implemented using the tri-channel DCT-perceptron layers, contains 22.64\% fewer parameters than the baseline ResNet-20 model, but it even reaches a 0.09\% higher accuracy. 1C-HT-ResNet-20 and 3C-HT-ResNet-20 have the same parameter savings (44.39\% and 26.64\%) as 1C-DCT-ResNet-20 and 3C-DCT-ResNet-20, and their accuracy losses are both less than 0.6\% compared to the baseline model. The reason that the HT models get lower accuracy results than the corresponding DCT models is that HT is a binary transform, which means its ability for frequency information extraction is weaker than the DCT. As compensation, fewer computational cost is required to compute those HT models. Fig.~\ref{fig: CIFAR-10} shows the test error history of the ResNet-20, 3C-HT-ResNet-20, and 3C-DCT-ResNet-20 on the CIFAR-10 test dataset.

Compared to our previous HT-based work in~\cite{pan2022block} which is also single-channel, the proposed 1C-HT-ResNet-20 reaches a 1.03\% higher accuracy. This is because the HT layer in~\cite{pan2022block} is channel-agnostic, while the proposed layer is channel-specific. Therefore, the proposed layers can utilize channel-wise information.

Another application for the proposed layers is that we can insert one additional layer before the global average pooling layer in the ResNets (the location is after the output of the final convolutional block). For example, an extra DCT-perceptron layer improves the accuracy of the regular ResNet-20 to 91.82\%. This method is named ResNet-20+1C-DCT-P.

\begin{table}[htbp]
\caption{CIFAR Ablation Study.}
\centering
\begin{tabular}{lcc}
	\hline\noalign{\smallskip}
	\bf{Method}&\bf{Parameters}&\bf{Accuracy}\\
	\noalign{\smallskip}\hline\noalign{\smallskip}
	\bf{1C-DCT-ResNet-20}&\bf{151,514}&\bf{91.59\%}\\
	(w/o shortcut in DCT-P.)&151,514&91.12\%\\
	(w/o scaling in DCT-P.)&147,482&90.46\%\\
	(ReLU w/ thresholds in DCT-P.)&151,514&91.30\%\\
	(ReLU w/o thresholds in DCT-P.)&147,818&91.06\%\\
	(Leaky ReLU w/ thresholds in DCT-P.)&151,514&90.79\%\\
	(SiLU w/ thresholds in DCT-P.)&151,514&91.39\%\\
	(replacing all Conv2D)&51,034&85.74\%\\
	\noalign{\smallskip}\hline\noalign{\smallskip}
	2C-DCT-ResNet-20&175,706&91.42\%\\
	\bf{3C-DCT-ResNet-20}&\bf{199,898}&\bf{91.75\%}\\
	(w/ shortcut in DCT-P.)&199,898&91.50\%\\
	4C-DCT-ResNet-20&224,090&91.48\%\\
	(w/ shortcut in DCT-P.)&224,090&91.46\%\\
	5C-DCT-ResNet-20&248,282&91.47\%\\
	(w/ shortcut in DCT-P.)&248,282&91.46\%\\
	\noalign{\smallskip}\hline\noalign{\smallskip}
	2C-HT-ResNet-20&175,706&91.28\%\\
	\bf{3C-HT-ResNet-20}&\bf{199,898}&\bf{91.29\%}\\
	4C-HT-ResNet-20&224,090&91.50\%\\
	5C-HT-ResNet-20&248,282&91.58\%\\
	6C-HT-ResNet-20&272,474&91.21\%\\
	\noalign{\smallskip}\hline\noalign{\smallskip}
	\bf{3C-DCT-ResNet-50}&\bf{20,742,020}&\bf{78.57\%}\\
	(w/o scaling in DCT-P.)&20,736,788&78.07\%\\
	(ReLU w/ thresholds in DCT-P.)&20,742,020&78.02\%\\
	(Leaky ReLU w/ thresholds in DCT-P.)&20,742,020&78.31\%\\
	(SiLU w/ thresholds in DCT-P.)&20,742,020&78.26\%\\
	\noalign{\smallskip}\hline\noalign{\smallskip}
	\multicolumn{3}{p{0.98\linewidth}}{Note: DCT-P. stands for the DCT-perceptron layer. ``w/" and ``w/o" are the abbreviations of ``with" and ``without". ResNet-20's revisions are evaluated on CIFAR-10, and ResNet-50's revisions are evaluated on CIFAR-100.}
\end{tabular}
\label{tab: CIFAR-10 ablation study}
\end{table}

Table~\ref{tab: CIFAR-10 ablation study} presents our ablation study on the CIFAR-10 dataset. First, we remove the residual design (the shortcut connection) in the single-channel DCT-perceptron layers and we obtain a worse accuracy (91.12\%). Therefore, the shortcut connection can improve the performance of the single-channel DCT-perceptron layer. Then, we remove scaling which is the convolutional filtering in the single-channel DCT-perceptron layers. In this case, the multiplication operations in the single-channel DCT-perceptron layer are only implemented in the $1\times 1$ Conv2D. The accuracy drops from 91.59\% to 90.46\%. Therefore, scaling is necessary to maintain accuracy, as it makes use of the convolution theorems. Next, we apply ReLU instead of soft-thresholding in the single-channel DCT-perceptron layers. We first apply the same thresholds as the soft-thresholding thresholds on ReLU. In this case, the number of parameters is the same as in the proposed 1C-DCT-ResNet-20 model, but the accuracy drops to 91.30\%. We then try regular ReLU with a bias term in the $1\times 1$ Conv2D layers in the single-channel DCT-perceptron layers, and the accuracy drops further to 91.06\%. These experiments show that the soft-thresholding is superior to the ReLU in this work, as the soft-thresholding retains negative frequency components with high amplitudes, which are also used in denoising and image coding applications. We also try the leaky ReLU function with a negative slope of 0.01 and SILU, but their performance is still worse than the soft-thresholding function.
Furthermore, we replace all Conv2D layers in the ResNet-20 model with the single-channel DCT-perceptron layers. We implement the downsampling (Conv2D with the stride of 2) by truncating the IDCT2D so that each dimension of the reconstructed image is one-half the length of the original. In this method, 81.31\% parameters are reduced, but the accuracy drops to 85.74\% due to lacking parameters.

In the ablation study on the multi-channel DCT-perceptron layer, we further try the structures with 2/4/5 channels, but their accuracy results are all lower than the tri-channel. A similar ablation study is taken on the HT-perceptron layer, and the best accuracy result is obtained using 5 channels. Normally, the more channels the layer has, the better performance it can reach because of the more parameters. However, too many channels may cause redundancy and make it hard to train the layer well using the back-propagation algorithm.

Furthermore, we add the shortcut connection in the 3/4/5-channel DCT-perception layers, but our experiments show that the shortcut connection is redundant in these DCT-perception layers. This may be because multi-channels are replacing the shortcut. The multi-channel structures allow the derivatives to propagate to earlier channels better than the single-channel structure. 

\subsection{CIFAR-100 Experiments}
We further evaluate the proposed methods using ResNet-50 on the CIFAR-100 dataset. To train ResNet-50 and its revision, we follow the hyper-parameter settings in~\cite{zhong2020random}. In detail, we use an SGD optimizer with a weight decay of 0.0005 and momentum of 0.9. The models are trained with a mini-batch size of 128 for 300 epochs. The initial learning rate is 0.1, and the learning rate is reduced by a factor of 1/10 at the 150th, and 225th epochs, respectively. We normalize the images with the mean of [0.5071, 0.4865, 0.4409] and the standard variations of [0.2673, 0.2564, 0.2762]. Other data augmentations are the same as in the CIFAR-10 experiments. Table~\ref{tab: CIFAR-100} presents the experiments, which show the tr-channel perceptron layers improve the accuracy and reduce the parameters and MACs for the ResNet-50. The ablation study to show that the soft-thresholding function is superior to ReLU and its variants are presented in Table~\ref{tab: CIFAR-10 ablation study}.

\begin{table*}[t]
\caption{CIFAR-100 Experimental Results.}
\centering
\begin{tabular}{lcccc}
	\hline\noalign{\smallskip}
	\bf{Method}&\bf{Parameters}&\bf{MACs (G)}&\bf{Top-1 Acc}&\bf{Top-5 Acc}\\
	\noalign{\smallskip}\hline\noalign{\smallskip}
	ResNet-50&23,705,252&1.31&78.04\%&94.17\%\\
	\noalign{\smallskip}\hline\noalign{\smallskip}
	1C-DCT-ResNet-50&19,743,812 (16.71\%$\downarrow$)&1.06 (18.8\%$\downarrow$)&77.40\%&93.84\%\\
	3C-DCT-ResNet-50&20,742,020 (12.50\%$\downarrow$)&1.13 (13.6\%$\downarrow$)&78.57\%&94.94\%\\
	1C-HT-ResNet-50&19,743,812 (16.71\%$\downarrow$)&1.04 (20.5\%$\downarrow$)&77.67\%&94.03\%\\
	3C-HT-ResNet-50&20,742,020 (12.50\%$\downarrow$)&1.11 (15.4\%$\downarrow$)&78.48\%&93.81\%\\ 
	1C-BWT-ResNet-50&19,743,812 (16.71\%$\downarrow$)&1.06 (18.8\%$\downarrow$)&77.81\%&94.33\%\\
	3C-BWT-ResNet-50&20,742,020 (12.50\%$\downarrow$)&1.13 (13.6\%$\downarrow$)&\bf{79.03\%}&\bf{95.06\%}\\
	\noalign{\smallskip}\hline\noalign{\smallskip}
\end{tabular}
\label{tab: CIFAR-100}
\end{table*}
\color{black}

\begin{table*}[t]
\caption{ImageNet-1K results. Center-crop top-1 accuracy is used for evaluation. The image size is $224\times 224$.}
\centering
\begin{tabular}{lccccccc}
	\hline\noalign{\smallskip}
	\multirow{2}{*}{\bf{Method}}&\multirow{2}{*}{\bf{Publication}}&\multirow{2}{*}{\bf{Parameters (M)}}&\multirow{2}{*}{\bf{MACs (G)}}&\multicolumn{2}{c}{\bf{Center-Crop Accuracy}}&\multicolumn{2}{c}{\bf{10-Crop Accuracy}}\\
	&&&&\bf{Top-1}&\bf{Top-5}&\bf{Top-1}&\bf{Top-5}\\
	\noalign{\smallskip}\hline\noalign{\smallskip}
	ResNet-18 \cite{he2016deep} (Torchvision~\cite{torchvision}, baseline)&CVPR2015&11.69&1.82&69.76\%&89.08\%&71.86\%&90.60\%\\
	DCT-based ResNet-18~\cite{xu2021dct}&IJCNN2021&8.74&-&68.31\%&88.22\%&-&-\\
	1C-DCT-ResNet-18&This Work&6.14 (47.5\%$\downarrow$)&1.47 (19.3\%$\downarrow$)&67.84\%&87.73\%&70.09\%&89.40\%\\
	3C-DCT-ResNet-18&This Work&7.56 (35.3\%$\downarrow$)&1.57 (13.6\%$\downarrow$)&69.55\%&89.04\%&71.91\%&90.55\%\\
	3C-DCT-ResNet-18+1C-DCT-P.&This Work&7.82 (33.1\%$\downarrow$)&1.59 (12.8\%$\downarrow$)&70.00\%&89.07\%&72.33\%&90.54\%\\
	\noalign{\smallskip}\hline\noalign{\smallskip}
	ResNet-50 \cite{he2016deep} (Torchvision~\cite{torchvision})&CVPR2015&25.56&4.12&76.13\%&92.86\%&77.43\%&93.75\%\\
	Order 1 + ScatResNet-50~\cite{oyallon2018compressing}&ECCV2018&27.8&-&74.5\%&92.0\%&-&-\\
	JPEG-ResNet-50~\cite{gueguen2018faster}&NIPS2018 &28.4&5.4&76.06\%&93.02\%&-&-\\
	ResNet-50+DCT+FBS (3x32)~\cite{dos2020good}&ICIP2020 &26.2&3.68&70.22\%&-&-&-\\
	ResNet-50+DCT+FBS (3x16)~\cite{dos2020good}&ICIP2020 &25.6&3.18&67.03\%&-&-&-\\
	
	FFC-ResNet-50 (+LFU, $\alpha=0.25$)~\cite{chi2020fast}&NIPS2020&26.7&4.3$^*$&77.8\%&-&-&-\\
	FFC-ResNet-50 ($\alpha=1$)~\cite{chi2020fast}&NIPS 2020&34.2&5.6$^*$&75.2\%&-&-&-\\
	Faster JPEG-ResNet-50~\cite{dos2021less}&CIARP2021 &25.1&2.86&70.49\%&-&-&-\\
	DCT-based ResNet-50~\cite{xu2021dct}&IJCNN2021&21.30&-&74.73\%&92.30\%&-&-\\
	
	ResNet-50 (AugSkip) ZerO Init~\cite{zhao2021zero}&ICLR2022&25.56&4.12&76.37\%&-&-&-\\
	ResNet-50 (our trial, baseline)&-&25.56&4.12&76.06\%&92.85\%&77.53\%&93.75\%\\
	3C-DCT-ResNet-50&This Work&22.62 (11.5\%$\downarrow$)&3.63 (12.0\%$\downarrow$)&76.56\%&93.15\%&78.04\%&94.05\%\\
	3C-HT-ResNet-50&This Work&22.63 (11.5\%$\downarrow$)&3.60 (12.6\%$\downarrow$)&76.36\%&93.02\%&77.79\%&94.02\%\\
	3C-BWT-ResNet-50&This Work&22.63 (11.5\%$\downarrow$)&3.75 (9.0\%$\downarrow$)&76.53\%&93.13\%&77.98\%&94.11\%\\
	\noalign{\smallskip}\hline\noalign{\smallskip}
	Harm-DCT-ResNet-50~\cite{ulicny2022harmonic} (100 epochs)&PR2022&25.6&-&77.02\%&93.36\%&-&-\\
	Harm-DCT-ResNet-50, progr.$\lambda$~\cite{ulicny2022harmonic} (100 epochs)&PR2022&19.7&-&76.79\%&93.33\%&-&-\\
	3C-DCT-ResNet-50 (100 epochs)&This Paper&22.62 (11.5\%$\downarrow$)&3.63 (12.0\%$\downarrow$)&77.02\%&93.36\%&78.54\%&94.21\%\\
	3C-HT-ResNet-50 (100 epochs)&This Work&22.63 (11.5\%$\downarrow$)&3.60 (12.6\%$\downarrow$)&76.94\%&93.20\%&78.62\%&94.27\%\\
	3C-BWT-ResNet-50 (100 epochs)&This Paper&22.63 (11.5\%$\downarrow$)&3.75 (9.0\%$\downarrow$)&77.03\%&93.36\%&78.83\%&94.22\%\\
	\noalign{\smallskip}\hline\noalign{\smallskip}
	ResNet-50 (our trial, baseline, input $256^2$)&-&25.56&5.38&76.18\%&92.94\%&77.61\%&93.88\%\\
	3C-DCT-ResNet-50 (input $256^2$)&This Work&22.63 (11.5\%$\downarrow$)&4.76 (11.5\%$\downarrow$)&77.00\%&93.52\%&78.62\%&94.36\%\\
	
	3C-HT-ResNet-50 (input $256^2$)&This Work&22.63 (11.5\%$\downarrow$)&4.58 (14.9\%$\downarrow$)&76.77\%&93.26\%&78.33\%&94.14\%\\
	
	3C-BWT-ResNet-50 (input $256^2$)&This Work&22.63 (11.5\%$\downarrow$)&4.76 (11.5\%$\downarrow$)&77.07\%&93.43\%&78.45\%&94.34\%\\
	\noalign{\smallskip}\hline\noalign{\smallskip}
	ResNet-101 \cite{he2016deep} (Torchvision~\cite{torchvision})&CVPR2015&44.55&7.85&77.38\%&93.54\%&78.95\%&94.45\%\\ResNet-101 \cite{he2016deep} (our trial, baseline)&-&44.55&7.85&77.35\%&93.53\%&78.59\%&94.31\%\\
	3C-DCT-ResNet-101&This Work& 38.47 (13.6\%$\downarrow$)&6.76 (13.8\%$\downarrow$)&77.92\%&93.71\%&79.32\%&94.59\%\\
	3C-HT-ResNet-101&This Work& 38.48 (13.6\%$\downarrow$)&6.81 (13.2\%$\downarrow$)&77.72\%&93.87\%&79.25\%&94.61\%\\
	3C-BWT-ResNet-101&This Work& 38.48 (13.6\%$\downarrow$)& 6.99 (11.0\%$\downarrow$)&77.92\%&93.89\%&79.43\%&94.77\%\\
	\noalign{\smallskip}\hline\noalign{\smallskip}
	ResNet-101 \cite{he2016deep} (our trial, baseline, input $256^2$)&-&44.55&10.25&77.90\%&94.00\%&79.32\%&94.71\%\\
	3C-DCT-ResNet-101 (input $256^2$)&This Work& 38.48 (13.6\%$\downarrow$)&8.86 (13.6\%$\downarrow$)&78.35\%&94.12\%&79.73\%&94.88\%\\
	3C-HT-ResNet-101 (input $256^2$)&This Work& 38.48 (13.6\%$\downarrow$)& 8.65 (15.7\%$\downarrow$)&78.16\%&93.96\%&79.60\%&94.80\%\\
	3C-BWT-ResNet-101 (input $256^2$)&This Work& 38.48 (13.6\%$\downarrow$)&8.86 (13.6\%$\downarrow$)&78.44\%&94.06\%&79.76\%&94.85\%\\
	\noalign{\smallskip}\hline\noalign{\smallskip}
	\multicolumn{8}{p{0.97\textwidth}}{Note: MACs of DCT/BWT in this work are computed using the matrix-vector product. These numbers can be reduced further if the transforms are implemented using the Cooley–Tukey algorithm~\cite{vetterli1985fast}. $^*$MACs of the FFT in~\cite{chi2020fast} are omitted by its authors.} 
	\end{tabular}
	\label{tab: ImageNet-1K}
\end{table*}

\begin{table*}
\caption{ImageNet-1K results of inserting an additional DCT-perceptron layer. The image size is $224\times 224$.}
\centering
\begin{tabular}{lcccccc}
\hline\noalign{\smallskip}
\multirow{2}{*}{\bf{Method}}&\multirow{2}{*}{\bf{Parameters (M)}}&\multirow{2}{*}{\bf{MACs (G)}}&\multicolumn{2}{c}{\bf{Center-Crop Accuracy}}&\multicolumn{2}{c}{\bf{10-Crop Accuracy}}\\
&&&\bf{Top-1}&\bf{Top-5}&\bf{Top-1}&\bf{Top-5}\\
\noalign{\smallskip}\hline\noalign{\smallskip}
ResNet-18&11.69&1.82&69.76\%&89.08\%&71.86\%&90.60\%\\
\bf{ResNet-18+1C-DCT-P.}&\bf{11.95 (2.3\%$\uparrow$)}&\bf{1.84 (0.3\%$\uparrow$)}&\bf{70.50\%}&\bf{89.29\%}&\bf{72.89\%}&\bf{90.98\%}\\
ResNet-50&25.56&4.12&76.06\%&92.85\%&77.53\%&93.75\%\\
\bf{ResNet-50+1C-DCT-P.}&\bf{29.76 (16.4\%$\uparrow$)}&\bf{4.33 (5.1\%$\uparrow$)}&\bf{76.09\%}&\bf{93.80\%}&\bf{77.67\%}&\bf{93.84\%}\\
\noalign{\smallskip}\hline\noalign{\smallskip}
\end{tabular}
\label{tab: ImageNet-1K extra}
\end{table*}

\subsection{ImageNet-1K Classification}
In this section, we employ PyTorch's official ImageNet-1K training code~\cite{ImageNet_training_in_PyTorch}. Because we use the PyTorch official training code with the default training configuration for the revised ResNet-18s training, we use PyTorch Torchvision's officially trained ResNet-18 model as the baseline. On the other hand, the default training for ResNet-50 requires about 26--30 GB GPU memory, while an NVIDIA RTX3090 GPU only has 24GB. To overcome the issue of memory deficiency, we halve the batch size and the learning rate correspondingly. We use an SGD optimizer with a weight decay of 0.0001 and momentum of 0.9 for 90 epochs. Revised ResNet-18's are trained with the default setting: a mini-batch size of 256, and an initial learning rate of 0.1. ResNet-50 and its revised models are trained with a mini-batch size of 128, the initial learning rate is 0.05. The learning rate is reduced by a factor of 1/10 after every 30 epochs. For data argumentation, we apply random resized crops on training images to get 224 by 224 images, then we randomly flip images horizontally. In testing on the ImageNet-1K validation dataset, we resize the images to make the smaller edge of images 256, then apply center crops on the resized testing images to get 224 by 224 images. We normalize the images with the means of [0.485, 0.456, 0.406] and the standard variations of [0.229, 0.224, 0.225], respectively. We evaluate our models on the ImageNet-1K validation dataset and compare them with the state-of-art papers. During the training, the best models are saved based on the center-crop top-1 accuracy on the ImageNet-1K validation dataset, and their accuracy numbers are presented in Table~\ref{tab: ImageNet-1K}. Furthermore, we carry out experiments with the input size of 256 by 256, for 256 is the smallest integer power of 2 which is larger than 224. The training strategies except for the image size are the same as the experiments with the image size of 224 by 224. We resize the validation images to make the smaller edge 292, then apply center crops on the resized testing images to get 256 by 256 images.
In Table~\ref{tab: ImageNet-1K}, the input sizes in the last 4 rows are $256\times256$, and in other rows are $224\times 224$. 
Fig.~\ref{fig: ImageNet-1K} shows the test error history of ResNet-50 and our revised ResNet-50s on the ImageNet-1K validation dataset.

When the input size is 224 by 224, 3C-HT-ResNet-50 and 3C-BWT-ResNet-50 have more parameters and more MACs than 3C-DCT-ResNet-50. This is because the input size is not a pair of integers power of 2, so the hidden tensors' sizes in the HT-perceptron layer and the BWT-perceptron layer are larger than in the DCT-perceptron layer. Although the HT is still more efficient than the DCT, the scaling, $1\times 1$ Conv2D, and soft-thresholding in the HT-perceptron layers need to be computed with tensors with larger sizes. Furthermore, if the input size satisfies the constraint of integers power of 2, 3C-HT-ResNet-50 takes less computational cost than 3C-DCT-ResNet-50, and 3C-BWT-ResNet-50 takes the same computational cost as 3C-DCT-ResNet-50. As Table~\ref{tab: ImageNet-1K} presents, all transform-based perceptron layers manage to improve the accuracy for ResNet-50 by reducing the number of parameters and the MACs. For example, the 3C-BWT-ResNet layer improves the center-crop top-1 accuracy of ResNet-50 from 76.18\% to 77.07\%, while 11.5\% parameters and MACs are reduced.


Other state-of-art transform-based methods are listed in Table~\ref{tab: ImageNet-1K} for comparison. 
In FFC-ResNet-50 (+LFU, $\alpha=0.25$)~\cite{chi2020fast}, authors use their proposed Fourier Units (FU) and local Fourier units (LFU) to process only 25\% channels of each input tensor and use the regular $3\times 3$ Conv2D to process the remaining to get the accuracy of 77.8\%. Although this method improves the accuracy, it also increases the number of parameters and MACs. We can also use a combination of regular the $3\times 3$ Conv2D layers and our proposed layers to increase the accuracy by increasing the parameters in their method. However, according to the ablation study in~\cite{chi2020fast}, if all channels are processed by their proposed FU, FFC-ResNet-50 ($\alpha=1$) only gets the accuracy of 75.2\%. The reported number of parameters and MACs are much higher than the baseline in~\cite{chi2020fast} even though MACs from the FFT and IFFT are not counted. Our methods improve accuracy for regular ResNet-50s with significantly reduced parameters and MACs. Models in~\cite{ulicny2022harmonic} were trained with 10 additional epochs. Our results can also be improved with further training. With 10 additional epochs training, our 3C-DCT-ResNet-50 obtains the same accuracy as Harm-DCT-ResNet-50 but with fewer parameters and MACs, and our 3C-BWT-RsNet-50 obtains a slightly better accuracy.
Pruning-based methods~\cite{li2016pruning, molchanov2019importance} can be implemented as a part of our method to prune some layers because we implement convolutions in the transform domain. For example, removing convolutional filtering is equivalent to removing the corresponding DCT, scaling, soft-thresholding, and inverse DCT operation.

The experiments of inserting one additional layer before the global average pooling layer in the ResNet-18 and the ResNet-50 are presented in Table~\ref{tab: ImageNet-1K extra}. The additional single-channel DCT-perceptron layer improves the accuracy of ResNet-18 and ResNet-50 respectively with negligible extra parameters and computational cost.

\subsection{Wall-Clock Time}
In this section, we compare the actual inference time of the proposed transform domain ResNets with the vanilla ResNets.  We use the CUDA-based HT implementation described in \cite{thomas2018learning}. The DCT and BWT are implemented as a matrix-vector product without taking advantage of their fast $(O(N^2\log_2 N))$ algorithms~\cite{cetin1993block,vetterli1985fast} because PyTorch does not support DCT and the wavelet transforms at the moment, and we could not find any third-party PyTorch implementation with CUDA acceleration. Accelerating the HT-perceptron layers is critical in the work of reducing the computational time of its implementation. Here, CUDA-based implementation of the HT resulted in a significant impact on the speed. The speed test is performed on the NVIDIA RTX3090 GPU using the batch size of 1. The CPU of the computer is Intel(R) Xeon(R) W-2255 CPU @ 3.70GHz. We first warm up the device by 10 inference times, then compute the average inference time of 100 inferences. We repeat the experiments 10 times and present the minimum time. 
When the base width is 128, our 1C-HT-ResNet-50 (10.19 mS) becomes faster than the vanilla ResNet-50 (10.84 mS) for an input of size $3\times 224\times 224\rightarrow 1000$ as shown in Table~\ref{tab: wall-clock}. When the base width is 64 as in the original design, our 1C-HT-ResNet-50 (7.55 mS) is a little bit slower than the vanilla ResNet-50 (7.44 mS). 
Although our proposed layers have fewer MACs, they require more memory than the $3\times3$ Conv2D layer. Section~\ref{sec: Limitations and Future Research} will discuss memory usage in detail. The additional memory usage slows down the inference speed of the proposed models.
As the base width (the number of neurons in layer ``Conv1") of the ResNet increases, the inference latency growth speed of our models is slower than the vanilla models.
This is because as Table~\ref{tab: parameters and MACs} presents, the coefficient on the $N^2C^2$ MACs term of our three-channel perceptron layers is 3, while the $3\times 3$ Conv2D layer is 9.

\begin{table}[tb]
\caption{Wall-Clock Time.}
\centering
\begin{tabular}{>{\color{black}}l>{\color{black}}c>{\color{black}}c>{\color{black}}c>{\color{black}}c}
\hline\noalign{\smallskip}
&\multirow{2}{*}{\bf{Base}}&&\multicolumn{2}{c}{\bf{Time (mS)}}\\
\bf{Input/Output}&\multirow{2}{*}{\bf{Width}}&\bf{Method}&\bf{Pure}&\multirow{2}{*}{\bf{CUDA}}\\
&&&\bf{Python}&\\
\noalign{\smallskip}\hline\noalign{\smallskip}
&&ResNet-20&-&\bf{1.78}\\
$3\times 32^2\rightarrow 10$&16&1C-HT-ResNet-20&8.32&4.10\\
&&3C-HT-ResNet-20&10.20&6.61\\
\noalign{\smallskip}\hline\noalign{\smallskip}
&&ResNet-20&-&11.75\\
$3\times 32^2\rightarrow 10$&384&1C-HT-ResNet-20&11.77&\bf{8.91}\\
&&3C-HT-ResNet-20&13.26&10.94\\
\noalign{\smallskip}\hline\noalign{\smallskip}
&&ResNet-20&-&21.26\\
$3\times 32^2\rightarrow 10$&512&1C-HT-ResNet-20&15.64&\bf{14.17}\\
&&3C-HT-ResNet-20&18.20&16.56\\
\noalign{\smallskip}\hline\noalign{\smallskip}
&&ResNet-50&-&\bf{5.11}\\
$3\times 32^2\rightarrow 100$&64&1C-HT-ResNet-50&10.06&6.79\\
&&3C-HT-ResNet-50&10.82&8.77\\
\noalign{\smallskip}\hline\noalign{\smallskip}
&&ResNet-50&-&19.75\\
$3\times 32^2\rightarrow 100$&256&1C-HT-ResNet-50&18.76&\bf{17.49}\\
&&3C-HT-ResNet-50&19.90&18.69\\
\noalign{\smallskip}\hline\noalign{\smallskip}
&&ResNet-50&-&\bf{7.44}\\
$3\times 224^2\rightarrow 1000$&64&1C-HT-ResNet-50&10.29&7.55\\
&&3C-HT-ResNet-50&13.88&8.90\\
\noalign{\smallskip}\hline\noalign{\smallskip}
&&ResNet-50&-&10.84\\
$3\times 224^2\rightarrow 1000$&128&1C-HT-ResNet-50&14.23&\bf{10.19}\\
&&3C-HT-ResNet-50&14.68&11.39\\
\noalign{\smallskip}\hline\noalign{\smallskip}
&&ResNet-50&-&32.98\\
$3\times 224^2\rightarrow 1000$&256&1C-HT-ResNet-50&31.63&\bf{30.55}\\
&&3C-HT-ResNet-50&33.99&32.81\\
\noalign{\smallskip}\hline\noalign{\smallskip}
&&ResNet-50&-&\bf{8.09}\\
$3\times 256^2\rightarrow 1000$&64&1C-HT-ResNet-50&14.09&9.17\\
&&3C-HT-ResNet-50&14.99&10.29\\
\noalign{\smallskip}\hline\noalign{\smallskip}
\multicolumn{5}{p{0.98\linewidth}}{Note: ``CUDA" and ``Pure Python" represent whether CUDA acceleration is applied on the HT-Perceptron layers. Other layers are built in PyTorch so they are always accelerated by CUDA. The pure Python implementation implements the transform using matrix multiplication. Therefore, the speed of DCT-ResNets and BWT-ResNets is the same as in this implementation.}
\end{tabular}
\label{tab: wall-clock}
\end{table}

Furthermore, we compare the inference latency with other works. We employ the model files from their official GitHub. As Table~\ref{tab: Inference latency} presents, our proposed 1C-HT-ResNet-50 and 3C-HT-ResNet-50 have the comparable inference latency as ResNet-50~\cite{he2016deep} and Harm-DCT-ResNet-50~\cite{ulicny2022harmonic}. FFC-ResNet-50~\cite{chi2020fast} is significantly slower than other models as FFT is a complex-valued transform. 

\begin{table}[tb]
\caption{Inference latency on $3\times 224^2\rightarrow 1000$. The base width is 64.}
\centering
\begin{tabular}{lc}
\hline\noalign{\smallskip}
\bf{Methods}&\bf{Time (mS)}\\
\noalign{\smallskip}\hline\noalign{\smallskip}
ResNet-50~\cite{he2016deep}&7.44\\
FFC-ResNet-50 (+LFU, $\alpha = 0.25$)~\cite{chi2020fast}&39.13\\
FFC-ResNet-50 ($\alpha = 1$)~\cite{chi2020fast}&39.52\\
Harm-DCT-ResNet-50~\cite{ulicny2022harmonic}&7.71\\
1C-HT-ResNet-50&7.55\\
3C-HT-ResNet-50&8.90\\
\noalign{\smallskip}\hline\noalign{\smallskip}
\end{tabular}
\label{tab: Inference latency}
\end{table}


\section{Limitations and Future Research}\label{sec: Limitations and Future Research}
The first limitation is the input size. The HT-perceptron and the BWT-perceptron layers prefer the input width and heights to be integers power of 2, otherwise, zero padding is required. The second limitation is the memory usage. In the $P$-channel structures, the memory usage will increase $P$ times. In practice, when we run the PyTorch official ImageNet training code~\cite{ImageNet_training_in_PyTorch} with a batch size of 256 and the input size of $3\times 224\times224$, ResNet-50 takes about 29 GB GPU memory. On the other hand, 3C-DCT-ResNet-50 takes about 36 GB. Both 3C-HT-ResNet-50 and 3C-BWT-ResNet-50 take about 39 GB (they take more memory because DCT does not require the input size to be an integer power of 2, so DCT does not need zero padding). The single-channel structures still require more parameters than ResNet-50. 1C-DCT-ResNet-50 takes about 31 GB. Both 1C-HT-ResNet-50 and 1C-BWT-ResNet-50 take about 33 GB. This is because the proposed layers are implemented in multiple steps, while the Conv2D layer is well-optimized in PyTorch. Saving these hidden tensors for the backpropagation algorithm takes more memory than the Conv2D layer. Therefore, increased pressure on the memory subsystem slows down the revised ResNet models though MACs are reduced by the proposed layers.

In the future, we will extend the application to other computer-vision-based tasks such as segmentation and reconstruction. Another research topic is to use smaller blocks as in vision transformers~\cite{liu2021swin} to reduce memory usage and latency. 

\section{Conclusion}\label{sec: Conclusion}
In this paper, we proposed a set of novel layers based on the discrete cosine transform (DCT), the Hadamard transform (HT), and the biorthogonal wavelet transform (BWT) to replace some of the $3\times 3$ Conv2D layers in convolutional neural networks (CNNs). All of our transform domain layers can be implemented with single-channel or multi-channel structures. The key idea of this work is based on the convolution theorems, that the convolution in the time and the space domain is equivalent to the element-wise multiplication in the transform domain. With the proposed layers, CNNs such as ResNets can obtain comparable or even higher accuracy results in image classification tasks with much fewer parameters. ResNets with the HT-perceptron layers usually produce a little lower accuracy than ResNets with the DCT-perceptron layers. HT is a binary transform, so it cannot approximate the Fourier transform as well as the DCT. On the other hand,
HT can be implemented without performing any multiplications.
Another application for the proposed layers is the addition of an extra layer to the regular ResNets. An additional transform domain layer before the global average pooling layer improves the accuracy of ResNets with a negligible increase in the number of parameters and computational time. In addition to ResNets,
the proposed transform domain layers can be used in any convolutional neural network using the Conv2D layers. 

\section{Acknowledgement}
The authors would like to express our sincere gratitude to Prof. Ulas Bagci and his research group for generously providing NVIDIA A100 GPUs for conducting the ResNet-101-related experiments in this study. Their support was instrumental in enabling the computational aspects of our research, and their expertise greatly contributed to the success of our work. The preliminary results of the HT-perceptron layers for quantum computing were published in~\cite{pan2023hybrid}. 

\bibliographystyle{IEEEtran}
\bibliography{main}

\begin{thebibliography}{10}
\providecommand{\url}[1]{#1}
\csname url@samestyle\endcsname
\providecommand{\newblock}{\relax}
\providecommand{\bibinfo}[2]{#2}
\providecommand{\BIBentrySTDinterwordspacing}{\spaceskip=0pt\relax}
\providecommand{\BIBentryALTinterwordstretchfactor}{4}
\providecommand{\BIBentryALTinterwordspacing}{\spaceskip=\fontdimen2\font plus
\BIBentryALTinterwordstretchfactor\fontdimen3\font minus
  \fontdimen4\font\relax}
\providecommand{\BIBforeignlanguage}[2]{{%
\expandafter\ifx\csname l@#1\endcsname\relax
\typeout{** WARNING: IEEEtran.bst: No hyphenation pattern has been}%
\typeout{** loaded for the language `#1'. Using the pattern for}%
\typeout{** the default language instead.}%
\else
\language=\csname l@#1\endcsname
\fi
#2}}
\providecommand{\BIBdecl}{\relax}
\BIBdecl

\bibitem{he2016deep}
K.~He, X.~Zhang, S.~Ren, and J.~Sun, ``Deep residual learning for image
  recognition,'' in \emph{Proceedings of the IEEE conference on computer vision
  and pattern recognition}, 2016, pp. 770--778.

\bibitem{badawi2020computationally}
D.~Badawi, H.~Pan, S.~C. Cetin, and A.~E. {\c{C}}etin, ``Computationally
  efficient spatio-temporal dynamic texture recognition for volatile organic
  compound (voc) leakage detection in industrial plants,'' \emph{IEEE Journal
  of Selected Topics in Signal Processing}, vol.~14, no.~4, pp. 676--687, 2020.

\bibitem{agarwal2021coronet}
C.~Agarwal, S.~Khobahi, D.~Schonfeld, and M.~Soltanalian, ``Coronet: a deep
  network architecture for enhanced identification of covid-19 from chest x-ray
  images,'' in \emph{Medical Imaging 2021: Computer-Aided Diagnosis}, vol.
  11597.\hskip 1em plus 0.5em minus 0.4em\relax International Society for
  Optics and Photonics, 2021, p. 1159722.

\bibitem{redmon2016you}
J.~Redmon, S.~Divvala, R.~Girshick, and A.~Farhadi, ``You only look once:
  Unified, real-time object detection,'' in \emph{Proceedings of the IEEE
  conference on computer vision and pattern recognition}, 2016, pp. 779--788.

\bibitem{aslan2020deep}
S.~Aslan, U.~G{\"u}d{\"u}kbay, B.~U. T{\"o}reyin, and A.~E. {\c{C}}etin, ``Deep
  convolutional generative adversarial networks for flame detection in video,''
  in \emph{International Conference on Computational Collective
  Intelligence}.\hskip 1em plus 0.5em minus 0.4em\relax Springer, 2020, pp.
  807--815.

\bibitem{menchetti2019pain}
G.~Menchetti, Z.~Chen, D.~J. Wilkie, R.~Ansari, Y.~Yardimci, and A.~E.
  {\c{C}}etin, ``Pain detection from facial videos using two-stage deep
  learning,'' in \emph{2019 IEEE Global Conference on Signal and Information
  Processing (GlobalSIP)}.\hskip 1em plus 0.5em minus 0.4em\relax IEEE, 2019,
  pp. 1--5.

\bibitem{aslan2019early}
S.~Aslan, U.~G{\"u}d{\"u}kbay, B.~U. T{\"o}reyin, and A.~E. {\c{C}}etin,
  ``Early wildfire smoke detection based on motion-based geometric image
  transformation and deep convolutional generative adversarial networks,'' in
  \emph{ICASSP 2019-2019 IEEE International Conference on Acoustics, Speech and
  Signal Processing (ICASSP)}.\hskip 1em plus 0.5em minus 0.4em\relax IEEE,
  2019, pp. 8315--8319.

\bibitem{beratouglu2021vehicle}
M.~S. Berato{\u{g}}lu and B.~U. T{\"o}reyin, ``Vehicle license plate detector
  in compressed domain,'' \emph{IEEE Access}, vol.~9, pp. 95\,087--95\,096,
  2021.

\bibitem{yu2018bisenet}
C.~Yu, J.~Wang, C.~Peng, C.~Gao, G.~Yu, and N.~Sang, ``Bisenet: Bilateral
  segmentation network for real-time semantic segmentation,'' in
  \emph{Proceedings of the European conference on computer vision (ECCV)},
  2018, pp. 325--341.

\bibitem{huang2019ccnet}
Z.~Huang, X.~Wang, L.~Huang, C.~Huang, Y.~Wei, and W.~Liu, ``Ccnet: Criss-cross
  attention for semantic segmentation,'' in \emph{Proceedings of the IEEE/CVF
  International Conference on Computer Vision}, 2019, pp. 603--612.

\bibitem{long2015fully}
J.~Long, E.~Shelhamer, and T.~Darrell, ``Fully convolutional networks for
  semantic segmentation,'' in \emph{Proceedings of the IEEE conference on
  computer vision and pattern recognition}, 2015, pp. 3431--3440.

\bibitem{poudel2019fast}
R.~P. Poudel, S.~Liwicki, and R.~Cipolla, ``Fast-scnn: fast semantic
  segmentation network,'' \emph{arXiv preprint arXiv:1902.04502}, 2019.

\bibitem{jin2019fast}
Y.~Jin, W.~Hao, P.~Wang, and J.~Wang, ``Fast detection of traffic congestion
  from ultra-low frame rate image based on semantic segmentation,'' in
  \emph{2019 14th IEEE Conference on Industrial Electronics and Applications
  (ICIEA)}.\hskip 1em plus 0.5em minus 0.4em\relax IEEE, 2019, pp. 528--532.

\bibitem{dong2017short}
X.~Dong, L.~Qian, and L.~Huang, ``Short-term load forecasting in smart grid: A
  combined cnn and k-means clustering approach,'' in \emph{2017 IEEE
  international conference on big data and smart computing (BigComp)}.\hskip
  1em plus 0.5em minus 0.4em\relax IEEE, 2017, pp. 119--125.

\bibitem{koyuncu2022centroidal}
E.~Koyuncu, ``Centroidal clustering of noisy observations by using $ r $ th
  power distortion measures,'' \emph{IEEE Transactions on Neural Networks and
  Learning Systems}, 2022.

\bibitem{miao2022federated}
R.~Miao and E.~Koyuncu, ``Federated momentum contrastive clustering,''
  \emph{arXiv preprint arXiv:2206.05093}, 2022.

\bibitem{cetin1993block}
A.~E. Cetin, O.~N. Gerek, and S.~Ulukus, ``Block wavelet transforms for image
  coding,'' \emph{IEEE Transactions on Circuits and Systems for Video
  Technology}, vol.~3, no.~6, pp. 433--435, 1993.

\bibitem{hartley1942more}
R.~V. Hartley, ``A more symmetrical fourier analysis applied to transmission
  problems,'' \emph{Proceedings of the IRE}, vol.~30, no.~3, pp. 144--150,
  1942.

\bibitem{chi2020fast}
L.~Chi, B.~Jiang, and Y.~Mu, ``Fast fourier convolution,'' \emph{Advances in
  Neural Information Processing Systems}, vol.~33, pp. 4479--4488, 2020.

\bibitem{mohammad2021substitution}
U.~F. Mohammad and M.~Almekkawy, ``A substitution of convolutional layers by
  fft layers-a low computational cost version,'' in \emph{2021 IEEE
  International Ultrasonics Symposium (IUS)}.\hskip 1em plus 0.5em minus
  0.4em\relax IEEE, 2021, pp. 1--3.

\bibitem{rao2023gfnet}
Y.~Rao, W.~Zhao, Z.~Zhu, J.~Zhou, and J.~Lu, ``Gfnet: Global filter networks
  for visual recognition,'' \emph{IEEE Transactions on Pattern Analysis and
  Machine Intelligence}, 2023.

\bibitem{david1995denoising}
L.~David and J.~Donoho, ``Denoising by soft-thresholding,'' \emph{IEEE
  Transactions on information theory}, vol.~41, no.~3, pp. 613--627, 1995.

\bibitem{gueguen2018faster}
L.~Gueguen, A.~Sergeev, B.~Kadlec, R.~Liu, and J.~Yosinski, ``Faster neural
  networks straight from jpeg,'' \emph{Advances in Neural Information
  Processing Systems}, vol.~31, 2018.

\bibitem{dos2020good}
S.~F. dos Santos, N.~Sebe, and J.~Almeida, ``The good, the bad, and the ugly:
  Neural networks straight from jpeg,'' in \emph{2020 IEEE International
  Conference on Image Processing (ICIP)}.\hskip 1em plus 0.5em minus
  0.4em\relax IEEE, 2020, pp. 1896--1900.

\bibitem{dos2021less}
S.~F. dos Santos and J.~Almeida, ``Less is more: Accelerating faster neural
  networks straight from jpeg,'' in \emph{Iberoamerican Congress on Pattern
  Recognition}.\hskip 1em plus 0.5em minus 0.4em\relax Springer, 2021, pp.
  237--247.

\bibitem{xu2021dct}
Y.~Xu and H.~Nakayama, ``Dct-based fast spectral convolution for deep
  convolutional neural networks,'' in \emph{2021 International Joint Conference
  on Neural Networks (IJCNN)}.\hskip 1em plus 0.5em minus 0.4em\relax IEEE,
  2021, pp. 1--8.

\bibitem{ulicny2022harmonic}
M.~Ulicny, V.~A. Krylov, and R.~Dahyot, ``Harmonic convolutional networks based
  on discrete cosine transform,'' \emph{Pattern Recognition}, vol. 129, p.
  108707, 2022.

\bibitem{zhao2021zero}
J.~Zhao, F.~Sch{\"a}fer, and A.~Anandkumar, ``Zero initialization: Initializing
  residual networks with only zeros and ones,'' \emph{arXiv}, 2021.

\bibitem{deveci2018energy}
T.~C. Deveci, S.~Cakir, and A.~E. Cetin, ``Energy efficient hadamard neural
  networks,'' \emph{arXiv preprint arXiv:1805.05421}, 2018.

\bibitem{pan2021fast}
H.~Pan, D.~Badawi, and A.~E. Cetin, ``Fast walsh-hadamard transform and
  smooth-thresholding based binary layers in deep neural networks,'' in
  \emph{Proceedings of the IEEE/CVF Conference on Computer Vision and Pattern
  Recognition}, 2021, pp. 4650--4659.

\bibitem{pan2022block}
------, ``Block walsh--hadamard transform-based binary layers in deep neural
  networks,'' \emph{ACM Transactions on Embedded Computing Systems}, vol.~21,
  no.~6, pp. 1--25, 2022.

\bibitem{liu2019multi}
P.~Liu, H.~Zhang, W.~Lian, and W.~Zuo, ``Multi-level wavelet convolutional
  neural networks,'' \emph{IEEE Access}, vol.~7, pp. 74\,973--74\,985, 2019.

\bibitem{oyallon2018compressing}
E.~Oyallon, E.~Belilovsky, S.~Zagoruyko, and M.~Valko, ``Compressing the input
  for cnns with the first-order scattering transform,'' in \emph{Proceedings of
  the European Conference on Computer Vision (ECCV)}, 2018, pp. 301--316.

\bibitem{karakucs2020simulation}
O.~Karaku{\c{s}}, I.~Rizaev, and A.~Achim, ``A simulation study to evaluate the
  performance of the cauchy proximal operator in despeckling sar images of the
  sea surface,'' in \emph{IGARSS 2020-2020 IEEE International Geoscience and
  Remote Sensing Symposium}.\hskip 1em plus 0.5em minus 0.4em\relax IEEE, 2020,
  pp. 1568--1571.

\bibitem{badawi2021discrete}
D.~Badawi, A.~Agambayev, S.~Ozev, and A.~E. Cetin, ``Discrete cosine transform
  based causal convolutional neural network for drift compensation in chemical
  sensors,'' in \emph{ICASSP 2021-2021 IEEE International Conference on
  Acoustics, Speech and Signal Processing (ICASSP)}.\hskip 1em plus 0.5em minus
  0.4em\relax IEEE, 2021, pp. 8012--8016.

\bibitem{pan2022deep}
H.~Pan, D.~Badawi, C.~Chen, A.~Watts, E.~Koyuncu, and A.~E. Cetin, ``Deep
  neural network with walsh-hadamard transform layer for ember detection during
  a wildfire,'' in \emph{Proceedings of the IEEE/CVF Conference on Computer
  Vision and Pattern Recognition}, 2022, pp. 257--266.

\bibitem{ahmed1974discrete}
N.~Ahmed, T.~Natarajan, and K.~R. Rao, ``Discrete cosine transform,''
  \emph{IEEE transactions on Computers}, vol. 100, no.~1, pp. 90--93, 1974.

\bibitem{strang1999discrete}
G.~Strang, ``The discrete cosine transform,'' \emph{SIAM review}, vol.~41,
  no.~1, pp. 135--147, 1999.

\bibitem{singh2011jpeg}
P.~Singh, P.~Singh, and R.~K. Sharma, ``Jpeg image compression based on
  biorthogonal, coiflets and daubechies wavelet families,'' \emph{International
  Journal of Computer Applications}, vol.~13, no.~1, pp. 1--7, 2011.

\bibitem{fino1976unified}
B.~J. Fino and V.~R. Algazi, ``Unified matrix treatment of the fast
  walsh-hadamard transform,'' \emph{IEEE Transactions on Computers}, vol.~25,
  no.~11, pp. 1142--1146, 1976.

\bibitem{vetterli1985fast}
M.~Vetterli, ``Fast 2-d discrete cosine transform,'' in \emph{ICASSP'85. IEEE
  International Conference on Acoustics, Speech, and Signal Processing},
  vol.~10.\hskip 1em plus 0.5em minus 0.4em\relax IEEE, 1985, pp. 1538--1541.

\bibitem{shukla2022hybrid}
A.~Shukla and P.~Vedula, ``A hybrid classical-quantum algorithm for solution of
  nonlinear ordinary differential equations,'' \emph{Applied Mathematics and
  Computation}, p. 127708, 2022.

\bibitem{shen1998dct}
B.~Shen, I.~K. Sethi, and V.~Bhaskaran, ``Dct convolution and its application
  in compressed domain,'' \emph{IEEE transactions on circuits and systems for
  video technology}, vol.~8, no.~8, pp. 947--952, 1998.

\bibitem{uvsakova2002walsh}
A.~Usakova, J.~Kotuliakov{\'a}, and M.~ZAJAC, ``Walsh--hadamard transformation
  of a convolution,'' \emph{Radioengineering}, vol.~11, no.~3, pp. 40--42,
  2002.

\bibitem{hendrycks2016gaussian}
D.~Hendrycks and K.~Gimpel, ``Gaussian error linear units (gelus),''
  \emph{arXiv preprint arXiv:1606.08415}, 2016.

\bibitem{javid2022developing}
I.~Javid, R.~Ghazali, I.~Syed, N.~A. Husaini, and M.~Zulqarnain, ``Developing
  novel t-swish activation function in deep learning,'' in \emph{2022
  International Conference on IT and Industrial Technologies (ICIT)}.\hskip 1em
  plus 0.5em minus 0.4em\relax IEEE, 2022, pp. 1--7.

\bibitem{zhong2020random}
Z.~Zhong, L.~Zheng, G.~Kang, S.~Li, and Y.~Yang, ``Random erasing data
  augmentation,'' in \emph{Proceedings of the AAAI conference on artificial
  intelligence}, vol.~34, no.~07, 2020, pp. 13\,001--13\,008.

\bibitem{torchvision}
``Models and pre-trained weights,''
  \url{https://pytorch.org/vision/stable/models.html}, 2022, accessed:
  2022-09-27.

\bibitem{ImageNet_training_in_PyTorch}
``Imagenet training in pytorch,''
  \url{https://github.com/pytorch/examples/tree/main/imagenet}, 2022, accessed:
  2022-09-27.

\bibitem{li2016pruning}
H.~Li, A.~Kadav, I.~Durdanovic, H.~Samet, and H.~P. Graf, ``Pruning filters for
  efficient convnets,'' \emph{arXiv preprint arXiv:1608.08710}, 2016.

\bibitem{molchanov2019importance}
P.~Molchanov, A.~Mallya, S.~Tyree, I.~Frosio, and J.~Kautz, ``Importance
  estimation for neural network pruning,'' in \emph{Proceedings of the IEEE/CVF
  conference on computer vision and pattern recognition}, 2019, pp.
  11\,264--11\,272.

\bibitem{thomas2018learning}
A.~Thomas, A.~Gu, T.~Dao, A.~Rudra, and C.~R{\'e}, ``Learning compressed
  transforms with low displacement rank,'' \emph{Advances in neural information
  processing systems}, vol.~31, 2018.

\bibitem{liu2021swin}
Z.~Liu, Y.~Lin, Y.~Cao, H.~Hu, Y.~Wei, Z.~Zhang, S.~Lin, and B.~Guo, ``Swin
  transformer: Hierarchical vision transformer using shifted windows,'' in
  \emph{Proceedings of the IEEE/CVF international conference on computer
  vision}, 2021, pp. 10\,012--10\,022.

\bibitem{pan2023hybrid}
H.~Pan, X.~Zhu, S.~F. Atici, and A.~Cetin, ``A hybrid quantum-classical
  approach based on the hadamard transform for the convolutional layer,'' in
  \emph{International Conference on Machine Learning}.\hskip 1em plus 0.5em
  minus 0.4em\relax PMLR, 2023, pp. 26\,891--26\,903.

\end{thebibliography}

\end{document}